\pdfoutput=1

\documentclass[11pt]{article}

\usepackage[]{eacl}

\usepackage{times}
\usepackage{latexsym}

\usepackage[T1]{fontenc}

\usepackage[utf8]{inputenc}

\usepackage{microtype}

\usepackage{inconsolata}

\usepackage{amssymb}
\usepackage{amsmath}
\usepackage{tikz}
\usepackage{verbatim}
\usepackage{multirow}
\usepackage{algorithm}
\usepackage{algorithmic}
\usepackage{subfigure}
\usepackage{graphicx}
\usepackage{makecell}
\usepackage{booktabs}
\usepackage{array}
\usepackage{arydshln}
\usepackage{booktabs}               

\DeclareMathOperator*{\argmax}{arg\,max}

\title{Graph-based Clustering for Detecting Semantic Change \\Across Time and Languages}

\author{Xianghe Ma$^{1}$ \text{} Michael Strube$^{2}$ \text{}
Wei Zhao$^{2 3}$ \\[0.4em]
    $^{1}$Faculty of Mathematics and Computer Science, Heidelberg University \\
    $^2$Heidelberg Institute for Theoretical Studies \\ 
    $^3$Department of Computing Science, University of Aberdeen \\
    \texttt{xianghe.ma@stud.uni-heidelberg.de \, michael.strube@h-its.org}\\
    \texttt{wei.zhao@abdn.ac.uk}\\
  }

\begin{document}
\maketitle
\begin{abstract}
Despite the predominance of contextualized embeddings in NLP,
approaches to detect semantic change relying on these embeddings and clustering methods underperform simpler counterparts based on static word embeddings. This stems from the poor quality of the clustering methods to produce sense clusters---which struggle to capture word senses, especially those with low frequency. This issue hinders the next step in examining how changes in word senses in one language influence another.
To address this issue, we propose a graph-based clustering approach to capture nuanced changes in both high- and low-frequency word senses across time and languages, including the acquisition and loss of these senses over time.
Our experimental results show that our approach substantially surpasses previous approaches in the SemEval2020 binary classification task across four languages.
Moreover, we showcase the ability of our approach as a versatile visualization tool to detect semantic changes in both intra-language and inter-language setups. We make our code and data publicly available\footnote{\url{https://gitlab.com/xiaohaima/lexical-dynamic-graph/}}.

\end{abstract}

\section{Introduction}
Since the 19th century, language change has been of  ongoing scholarly interest in historical linguistics, stemming from a curiosity to understand intricate genealogy of languages through the comparison of linguistic patterns across earlier and later text corpora \cite{bopp1816conjugationssystem, rask1818undersogelse, whitney1892max}. Up to now, many more curiosities have emerged, including the establishment of empirical principles of language change \cite{weinreich1968empirical, labov1972some, labov1982building, labov1994principles, labov2010principles}, the justification of hypothetical pathways of language change \cite{roberts2012diachrony, breitbarth201410, lehmann2015thoughts, breitbarth2019should}, the investigation of ancestral relationships among hundreds of languages \cite{boas1929classification, jager2013phylogenetic, guldemann2018historical}, the discovery of linguistic and extralinguistic factors driving language change \cite{blaxter2015gender}, etc.

\begin{figure}[htb]
\centering
\includegraphics[width=0.95\linewidth]{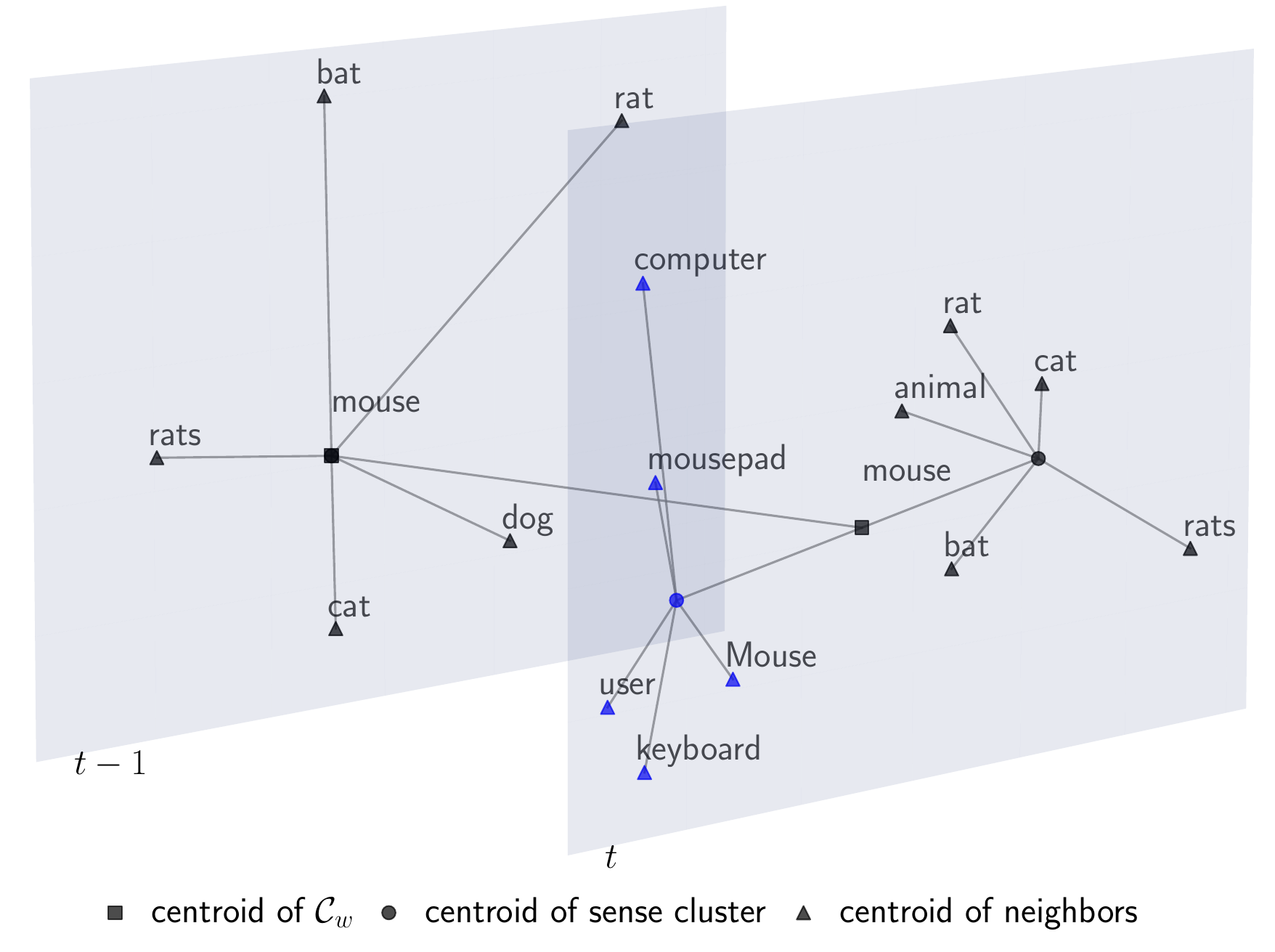}
\caption{Representation of the semantic changes for `mouse' in our temporal dynamic graph. Blue nodes indicate the acquisition of a new meaning over time, while black nodes indicate unchanged word meanings.
}
\label{fig:temporal-graph}
\end{figure}

A seminal work by \citet{coseriu1970einfuhrung} outlined the characteristics of language change along five dimensions: time, geographical places, medium, registers and social contexts. This has inspired many works to date that leverage these dimensions as a lens to examine changes in grammatical meaning \cite{closs1985regularity, bybee1985cross}, syntax \cite{hale1998diachronic, breitbarth2022v3} and many more.
In computational linguistics, there has been a surge of interest in leveraging machine learning methods, as cost-efficient alternatives to labor-intensive human inspection. A special focus has been given to detect lexical meaning change, aiming to track changes in word meanings through the analysis of word usages across different time periods \cite{rohrdantz-etal-2011-towards, eger-mehler-2016-linearity, hamilton-etal-2016-cultural, hamilton-etal-2016-diachronic, hamilton-etal-2016-diachronic, 
martinc2020capturing, gonen-etal-2020-simple, kaiser2021effects, 
montariol-etal-2021-scalable,teodorescu2022ualberta, d-zamora-reina-etal-2022-black}. 

For instance, \citet{prazak-etal-2020-uwb} and \citet{kaiser2021effects} leverage static word embeddings to represent target words across time periods, and then identify the presence of semantic change in each target word by assessing the similarity between its word embeddings from different time periods. \citet{kanjirangat-etal-2020-sst} and \citet{gyllensten2020sensecluster} employ contextualized word embeddings and a clustering method to detect changes in each word sense over time. Although contextualized embeddings excel in many NLP tasks, the performance of these embeddings coupled with clustering methods falls under static counterparts in detecting semantic change \cite{schlechtweg2020semeval}.

In this work, we identify two major limitations of previous works relying on contextualized embeddings and clustering methods, namely (a) they struggle to capture word senses, especially those with low frequency, leading to poor semantic representation of word senses and (b) they produce time-independent sense clusters and use them to represent word senses varying over time---which is particularly problematic in the presence of a big time gap (e.g., 100 years) between earlier and later time periods.
Moreover, these works are limited in scope to detect only intra-language semantic changes. To address these issues, we introduce a graph-based clustering approach that leverages contextualized embeddings to capture the evolution of each word sense across both time and languages. As a result, our approach allows for comparing changes in each word sense across languages over time. This allows for a detailed study of inter-language semantic change, especially to determine if the meanings of word translations across languages remain consistent or diverge over time.

We comparably evaluate our approach in the SemEval2020 binary classification and ranking tasks \cite{schlechtweg2020semeval} for detecting semantic change across English, German, Latin and Swedish, and investigate the potential of our graph-based approach, as a visualization tool, to detect semantic changes in both intra-language and inter-language setups. Our findings are summarized below:

\begin{itemize}
    \item Our approach substantially outperforms the SemEval2020 shared task winner \cite{prazak-etal-2020-uwb} in binary classification across four languages. Our ablation results demonstrate the effectiveness of three crucial components in our approach: our clustering strategy and method, and our distance metric. In the ranking task, our approach performs best in English but falls short in other languages compared to static embedding counterparts.
    
    \item We showcase the ability of our approach, as a versatile visualization tool, specifically to 
    (a) track nuanced intra-language semantic changes over time, including both the acquisition and loss of each word sense 
    and (b) track the consistency and divergence of semantic changes over time by comparing detected semantic changes within each language.
    This aids understanding of inter-language impacts on semantic changes, e.g., new meanings borrowed from other languages.
\end{itemize}

\section{Related Work}
\paragraph{Intra-Language Semantic Change Detection.} 
Recently, there has been a growing interest towards detecting meaning changes of target words within each language through a corpus-based study on word usage across time periods \cite{kutuzov2020uio, pomsl2020circe, giulianelli-etal-2020-analysing,
gyllensten2020sensecluster, 
karnysheva2020tue, kaiser2021effects,kutuzov-etal-2022-contextualized,card-2023-substitution}. Many approaches have been proposed in the SemEval2020 shared tasks \cite{schlechtweg2020semeval}. 
Most approaches fall under two categories, based on the choice of word embeddings. For \textbf{static embeddings}, approaches, such as \citet{prazak-etal-2020-uwb} and \citet{kaiser2021effects}, begin by refining pre-trained static word embeddings of target words on two corpora from different time periods, resulting in a separate embedding space for each time period. They then employ alignment techniques \cite{brychcin2019cross, artetxe-etal-2018-robust}
to adjust these word embeddings from different time periods. Lastly, a distance measure is applied to these adjusted word embeddings to detect semantic change. For \textbf{contextualized word embeddings}, approaches like \citet{kanjirangat-etal-2020-sst} and \citet{gyllensten2020sensecluster} employ the BERT and XLM-R encoders to produce contextualized word embeddings of each target word. 
They then employ $k$-means \cite{rousseeuw1987silhouettes} to partition embeddings of the target word from different time periods into multiple (time-independent) sense clusters. Lastly, a frequency-based criterion
is applied to these sense clusters to detect semantic change.
\citet{laicher-etal-2021-explaining} show that careful data preprocessing can further improve the performance of semantic change detection, and suggest encoding lemmatized target words instead of their original word forms. \citet{kutuzov-etal-2022-contextualized} propose to ensemble two top-performing approaches for detecting semantic changes.
\citet{kudisov-arefyev-2022-black} and \citet{card-2023-substitution} propose to detect semantic change by comparing two frequency distributions of a target word across time periods. Each distribution represents the frequencies of vocabulary words predicted as top substitutes for the target word using a masked language model.

Our work differs from others in several aspects: First, we leverage temporal and spatial dynamic graphs, which are derived from BERT embeddings, to represent changes in word meanings across time and space (languages)\footnote{Unlike \citet{schlechtweg-etal-2021-dwug}, which proposed human-annotated diachronic word usage graphs, our graphs are machine-generated through BERT and additionally offer visual clues regarding meaning changes over time.}. This allows for detecting nuanced changes in each word sense, including both the acquisition and loss of meanings over time. Second, we introduce our clustering method and strategy, and our distance metric, designed to produce sense clusters that excel in capturing word senses, especially for low-frequency senses. 
Moreover, we compute the similarity between sense clusters over time to detect semantic change, while previous approaches do so by applying a frequency-based criterion\footnote{Frequency-based criterion: Sense change is detected when a time-independent sense cluster has fewer than 2 tokens in the earlier corpus and more than 5 tokens in the later.}.

\paragraph{Inter-Language Semantic Change Detection.} 
While words in different languages may share a common ancestor and initial meaning, their meanings can diverge over time due to linguistic and extralinguistic variations in these languages. This intriguing phenomenon has led to increased research efforts towards studying semantic changes across languages. To do this, most previous works rely on semantic false friends, namely a pair of words in different languages that share an etymological origin but differ greatly in word meaning \cite{inkpen2005automatic, nakov2009unsupervised, chen2016false, st2017identifying, uban2019computational, uban2021cross}. For instance, 
\citet{uban2019computational} employ cross-lingual word embeddings to identify false friends, specifically to determine whether the current meanings of these word pairs have changed. \citet{uban2021cross} extended this idea by investigating cross-lingual semantic change laws, specifically by examining the meaning divergence of cognate words from their shared etymological origin. 

Furthermore, \citet{montariol2021measure} explored bilingual semantic divergence by comparing the meaning changes of mutual word translations of English and French using multilingual BERT. In contrast to our work, they only consider high-frequency word senses,
and do not differentiate between the acquisition and loss of meanings over time. Moreover, they do not provide a visual tool to detect semantic divergence across languages.

\section{Our Approach}
\subsection{Semantic-Tree Representation}
For each word $w$, let $\mathcal{C}_w = \{c_1, c_2, \dots, c_n\}$ be a word cloud consisting of a set of $d$-dimensional
contextualized word embeddings, where $n$ represents the word occurrence in a corpus.
We let $e_w \in \mathbb{R}^d$ denote the centroid of $\mathcal{C}_w$, given by $e_w = \frac{1}{n}\sum_{i}^{n}c_i$. For any two word clouds, the distance between their centroids $e_i$ and $e_j$ is denoted by $d(e_1, e_2) = 1 - \mathrm{sim}(e_1, e_2)$, where $\mathrm{sim}(e_1, e_2)$ is the cosine similarity between the centroids.

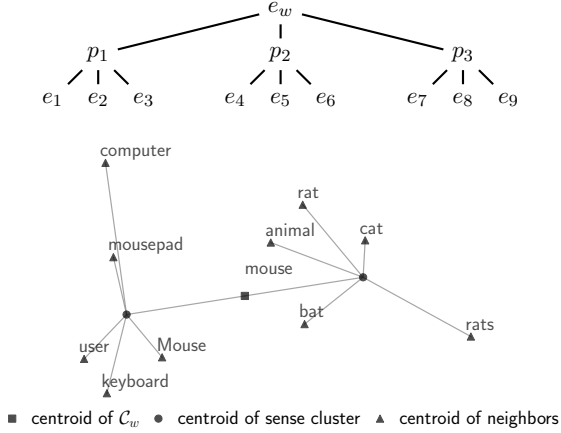
\begin{figure}[!tb]
\centering
\begin{subfigure}
    \centering
    \begin{tikzpicture}   
    [thick,scale=0.4, every node/.style={scale=0.8}]
    
    \node {$e_w$}
    child {node {$p_1$}
        child {node {$e_1$}}
        child {node {$e_2$}}
        child {node {$e_3$}}
    }    
    child [missing] {}    
    child [missing] {} 
    child [missing] {}        
    child { node {$p_2$}
        child {node {$e_4$}}
        child {node {$e_5$}}
        child {node {$e_6$}}
    }    
    child [missing] {}
    child [missing] {}    
    child [missing] {}    
    child { node {$p_3$}
        child {node {$e_7$}}
        child {node {$e_8$}}
        child {node {$e_9$}}
    };
\end{tikzpicture}
\end{subfigure}
\hfill
\hfill
\begin{subfigure}
    \centering 
\includegraphics[width=\linewidth]{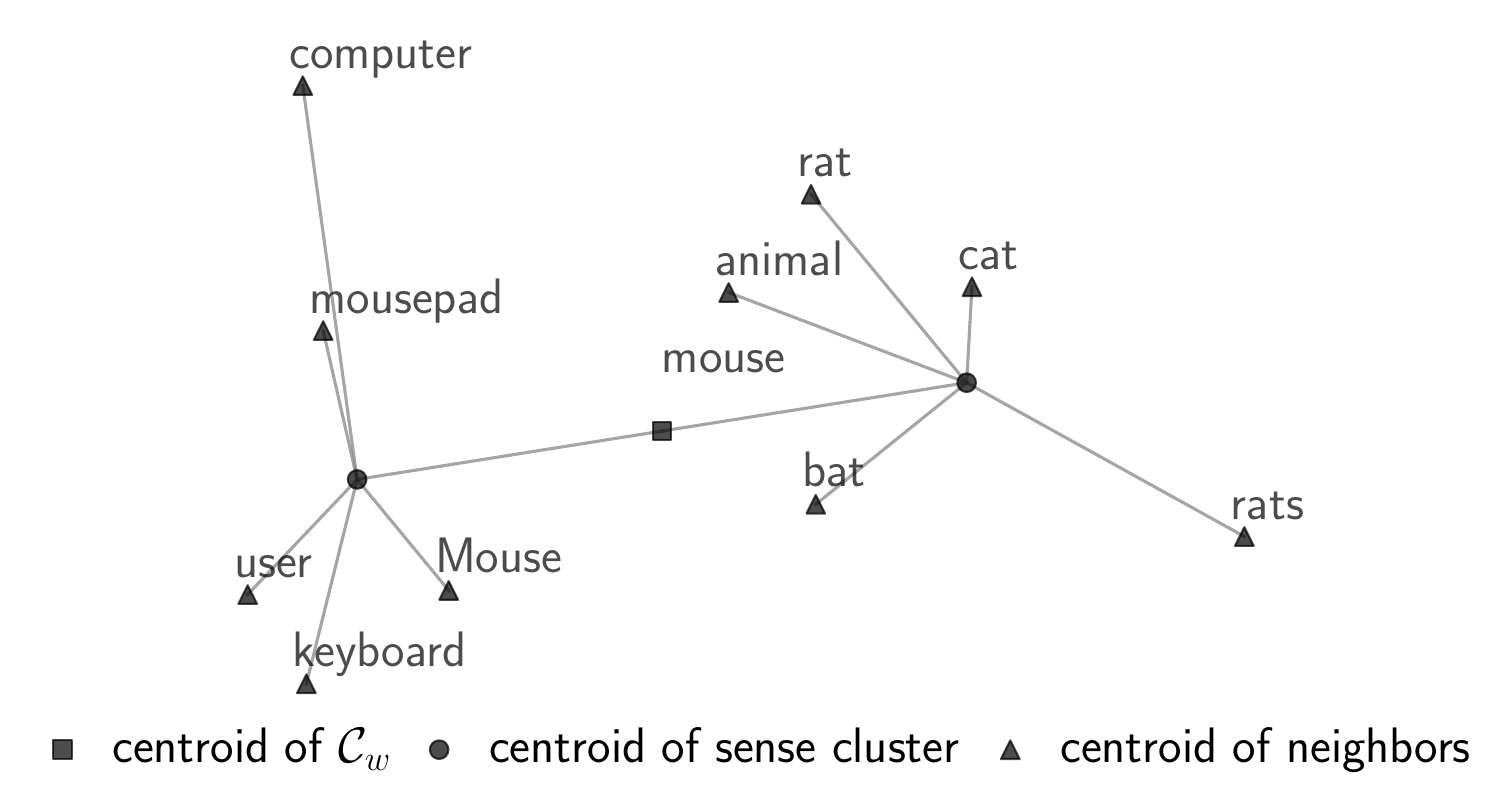}

\end{subfigure}
\caption{(Top) Representation of polysemous meanings of a word $w$ in a semantic-tree graph. (Bottom) Graph representation of `mouse' as the root node, generated by applying our approach to the English Wikipedia corpus. 
 \label{fig:tree-structure}}
\end{figure}

Each word $w$ may exhibit polysemy, manifesting different meanings depending on the context. Therefore, we partition $\mathcal{C}_w$ into $m$ sense clusters, i.e., $\mathcal{C}_w$ = $\bigcup_{1\leq i\leq m}\mathcal{C}_w (p_i)$. Each cluster $\mathcal{C}_w (p_i)$, which is a subset of $\mathcal{C}_w$ centered at $p_i$, represents a distinct meaning of the polysemous word. For each word, we let $\mathcal{P}_w=\{p_1, \dots, p_m\}$ denote a set of centroids corresponding to $m$ sense clusters. These centroids are determined using our clustering method (see \S\ref{sec:clustering}).
As illustrated in Figure \ref{fig:tree-structure}, we define a semantic-tree graph that captures multiple recorded meanings of a polysemous word $w$. We consider the root node $e_w$, the centroid of $\mathcal{C}_w$, as the \textbf{representative embedding}\footnote{We represent graph nodes as 2-dimensional embeddings, resulting from the PCA projection of high-dimensional contextualized word embeddings.} of the word $w$ reoccurring in a corpus. The root node is connected to three nodes on the second layer, which are three sense clusters' centroids $\{p_1, p_2, p_3\}$. 
We refer to the nodes on the third layer as the representative embeddings of three semantically nearest neighboring words to each centroid $p_i$.

\subsection{Temporal Dynamics within Semantics}
\label{sec:time-space}
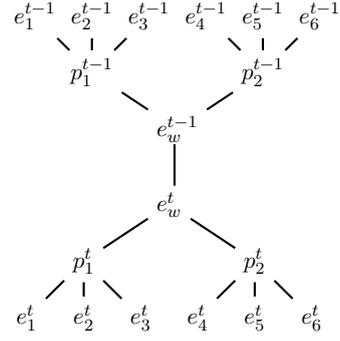
\begin{figure}
    \centering
\begin{tikzpicture}
[thick,scale=0.5, every node/.style={scale=0.8}]
\node [right=3cm]  at (0,-1)  {$e_w^{t-1}$}[rotate=180]
  child {node {$p_2^{t-1}$}
    child {node {$e_6^{t-1}$}}
    child {node {$e_5^{t-1}$}}
    child {node {$e_4^{t-1}$}}
    }
  child [missing] {}  
  child [missing] {}  
  child {node {$p_1^{t-1}$}
    child {node {$e_3^{t-1}$}}
    child {node {$e_2^{t-1}$}}
    child {node {$e_1^{t-1}$}}
    };

\draw (5.5,-1.4) -- (5.5,-2.5);

\node[right=3cm] at (0,-3) {$e_w^{t}$}
  child {node {$p_1^{t}$}
    child {node {$e_1^{t}$}}
    child {node {$e_2^{t}$}}
    child {node {$e_3^{t}$}}
    }
  child [missing] {}  
  child [missing] {}  
  child {node {$p_2^{t}$}
    child {node {$e_4^{t}$}}
    child {node {$e_5^{t}$}}
    child {node {$e_6^{t}$}}
    };
\end{tikzpicture}
    \caption{Representation of semantic changes over time in a temporal dynamic graph.
    }
    \label{fig:tree-structure-temporal}
\end{figure}

We add a temporal dimension to our semantic-tree graph for capturing meaning changes over time. To do this, we denote $\mathcal{C}_w^{t-1}$ and $\mathcal{C}_w^{t}$ as two point clouds of the word $w$ at two consecutive time periods $t-1$ and $t$. We then define a temporal dynamic graph to capture meaning shifts of the word $w$ over time,
as illustrated in Figure \ref{fig:tree-structure-temporal}. Given such a graph, we introduce our approach to detect changes in word meaning over time from $t-1$ to $t$ through a two-step process: (a) computing the similarity between sense clusters via our neighbor-based distance metric and (b) utilizing our detection criterion to detect the presence of semantic change.

\paragraph{Bipartite Matching Between Sense Clusters.} Our goal is to measure the similarity between a pair of sense clusters from different time periods, i.e., $C^{t-1}_w(p_i^{t-1})$ and $C^t_w(p_j^{t})$. This is achieved by measuring the similarity between the centroids of these sense clusters, i.e., $p_i^{t-1}$ and $p_j^{t}$, using a bipartite matching method and our \textbf{neighbor-based distance metric}\footnote{This metric leverages the participation of neighbors to determine the similarity between two words. By doing so, the similarity between two words is less affected by their embedding quality. See Figure \ref{fig:matching} (appendix) for the idea illustration.}.

To do this, we define $U_w = \{e_1^{t-1}, \dots, e_k^{t-1}\}$ as a set of the representative embeddings of $k$-nearest neighboring words to $p_i^{t-1}$ at time $t-1$. Each $e_i^{t-1}$ is the average of contextual word embeddings of a neighboring word. Similarly, we define $V_w = \{e_1^{t}, \dots, e_k^{t}\}$ as their counterparts to $p_j^{t}$ at time $t$. Our bipartite matching problem is given by:

\begin{small}
\begin{align*}
&\min_{\mu \in \{0,1\}^{k^2}} \sum_{e^{t-1} \in U_w} \sum_{e^{t} \in V_w} \mu (e^{t-1}, e^{t}) d(e^{t-1}, e^{t}) \\
\text{s.t.} 
& \sum_{e^{t} \in V_w} \mu (e^{t-1}, e^{t}) = 1, \quad \forall e^{t-1} \in U_w \\
& \sum_{e^{t-1} \in U_w} \mu (e^{t-1}, e^{t}) = 1, 
\quad \forall e^{t} \in V_w \\
& \mu (e^{t-1}, e^{t})  \in \{0, 1\}, \quad \forall e^{t-1} \in U_w, e^{t} \in V_w 
\end{align*}
\end{small}
where $\mu (e^{t-1}, e^{t}) $ denotes a binary variable that indicates whether a match exists between the input arguments, and $d(e^{t-1}, e^{t})$ represents the cosine distance between them. 
Lastly, the similarity between $p_i^{t-1}$ and $p_j^t$ is given by:

\begin{footnotesize}
\begin{align*}
s(p_i^{t-1}, p_j^t) = 1 - \frac{1}{|U_w|}\sum_{e^{t-1} \in U_w} \sum_{e^{t} \in V_w} \hat{\mu} (e^{t-1}, e^{t}) d(e^{t-1}, e^{t}) 
\end{align*}
\end{footnotesize}
where $\hat{\mu}$ is the optimal solution for bipartite matching, solved by using the Jonker-Volgenant algorithm \cite{crouse2016implementing}.

\paragraph{Semantic Change Detection.} 
Our goal is to identify meaning changes over time, especially to distinguish between the acquisition and loss of meanings. We now introduce our \textbf{detection criterion}:

For each word $w$, we denote $M_w=\{p_1^t, \dots, p_m^t \}$ as a set of the centroids of $m$ sense clusters at time $t$, and $N_w=\{p_1^{t-1}, \dots, p_n^{t-1}\}$ as counterparts of $n$ sense clusters at time $t-1$. We then compute the pairwise similarities between the two sets, yielding a semantic similarity matrix denoted below:

\begin{footnotesize}
\begin{align*}
\mathcal{S} = 
\begin{pmatrix}
s(p_1^{t-1}, p_1^t)&\dots&s(p_1^{t-1}, p_m^t)\\ 
\vdots&\ddots&\vdots\\ 
s(p_n^{t-1}, p_1^t)&\dots&s(p_n^{t-1}, p_m^t)\\ 
\end{pmatrix}
\end{align*}
\end{footnotesize}

Based on this matrix, we introduce a threshold $t_{sc}$ to differentiate between acquiring new meanings and losing existing ones:

\begin{itemize}
    \item If the word at time $t$ loses an existing meaning that it had at time $t-1$, namely $p_i^{t-1}$, then one cannot find any sense cluster centroids at time $t$ that are similar to $p_i^{t-1}$. This means that the similarity scores of $s(p_i^{t-1}, p_j^t)$ for all $j$ should fall below $t_{sc}$, i.e., all the entries in the $i$-th row of $\mathcal{S}$ are lower than $t_{sc}$.
    \item If a word gains a new meaning at time $t$, i.e., $p_j^{t}$, the $s(p_i^{t-1}, p_j^t)$ scores for all $i$ should fall below $t_{sc}$, i.e., all the entries in the $j$-th column of $\mathcal{S}$ are lower than $t_{sc}$. 
\end{itemize}
In either way, be it acquisition or loss of meanings, semantic change is detected.

\subsection{Temporal and Spatial Dynamics}
Here we extend our temporal dynamic graph to a cross-lingual setup, allowing us to detect semantic change across languages, especially to investigate whether the meanings of word translations across languages change consistently or diverge over time. To do this, we first introduce a spatial dynamic graph, and then combine it with a temporal dynamic graph for detecting/comparing semantic changes over time across languages.

\begin{figure}
    \centering
\begin{tikzpicture}
[thick,scale=0.5, every node/.style={scale=0.8}]
\node [right=3cm]  at (0,-1)  {$e_w^{\ell_1}$}[rotate=180]
  child {node {$p_2^{\ell_1}$}
    child {node {$e_6^{\ell_1}$}}
    child {node {$e_5^{\ell_1}$}}
    child {node {$e_4^{\ell_1}$}}
    }
  child [missing] {}  
  child [missing] {}  
  child {node {$p_1^{\ell_1}$}
    child {node {$e_3^{\ell_1}$}}
    child {node {$e_2^{\ell_1}$}}
    child {node {$e_1^{\ell_1}$}}
    };

\draw[densely dotted] (5.5,-1.4) -- (5.5,-2.5);

\node[right=3cm] at (0,-3) {$e_w^{\ell_2}$}
  child {node {$p_1^{\ell_2}$}
    child {node {$e_1^{\ell_2}$}}
    child {node {$e_2^{\ell_2}$}}
    child {node {$e_3^{\ell_2}$}}
    }
  child [missing] {}  
  child [missing] {}  
  child {node {$p_2^{\ell_2}$}
    child {node {$e_4^{\ell_2}$}}
    child {node {$e_5^{\ell_2}$}}
    child {node {$e_6^{\ell_2}$}}
    };
\end{tikzpicture}
    \caption{Representation of polysemous meanings of a mutual word translation pair in a spatial dynamic graph. 
    }
    \label{fig:tree-structure-spatial}
\end{figure}
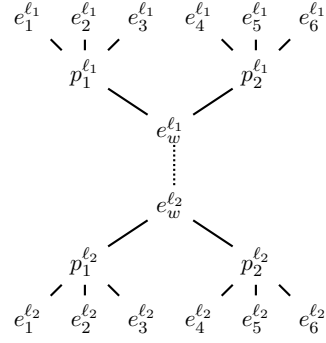

\paragraph{Spatial Dynamic Graph.} We let $w^{\ell_1}$ and $w^{\ell_2}$ be a pair of mutual word translations in languages $\ell_1$ and $\ell_2$. Denote $\mathcal{C}_w^{\ell_1}$ as a word cloud consisting of a set of the contextualized word embeddings of the word $w^{\ell_1}$, and $e^{\ell_1}_w$ as the word cloud's centroid, $\mathcal{C}_w^{\ell_1}(p_i^{\ell_1})$ as each sense cluster centered at $p_i^{\ell_1}$, and
$U_w^{\ell_1}=\{e_1^{\ell_1}, \dots, e_k^{\ell_1}\}$ as a set of the representative embeddings of $k$-nearest semantic neighboring words to $p_i^{\ell_1}$. Similarly, we denote $\mathcal{C}_w^{\ell_2}$, $e^{\ell_2}_w$, $p_j^{\ell_2}$, and $V_w^{\ell_2}=\{e_1^{\ell_2}, \dots, e_k^{\ell_2}\}$ as the counterparts in language $\ell_2$. Such a graph is depicted in Figure \ref{fig:tree-structure-spatial}.

We extend the idea of our bipartite matching method to a cross-lingual setup by leveraging $k$-nearest semantic neighbors $U_w^{\ell_1}$ and $V_w^{\ell_2}$ to compute the similarity between two sense clusters' centroids ($p_i^{\ell_1}$ and $p_j^{\ell_2}$) in different languages. 
Our bipartite matching problem is adjusted to:

\begin{small}
\begin{align*}
\min_{\mu \in \{0,1\}^{k^2}} \sum_{e^{\ell_1} \in U_w^{\ell_1}} \sum_{e^{\ell_2} \in V_w^{\ell_2}} \mu (e^{\ell_1}, e^{\ell_2}) d(e^{\ell_1} + b, e^{\ell_2})
\end{align*}
\end{small}
where $b$ is a rectified vector that addresses the misalignment between the embedding spaces of languages $\ell_1$ and $\ell_2$, assuming that one can translate one space to another by using this vector\footnote{We note that alignments would not affect the results of sense clusters in both source and target languages, as they only shift the embedding space of one language using a translation vector---which does not change the internal structure (topology) of the embedding space.}. We refer this vector to the difference between an average token embedding of all words in $\ell_1$ and its counterpart in $\ell_2$ \cite{liu2020study}. 

Once the optimal solution $\hat{u}$ is determined, the similarity between $p_i^{\ell_1}$ and $p_j^{\ell_2}$ is given by: 

\begin{small}
\begin{align*}
&s(p_i^{\ell_1}, p_j^{\ell_2}) = \\ 
&1 - \frac{1}{|U_w^{\ell_1}|}\sum_{e^{\ell_1} \in U_w} \sum_{e^{\ell_2} \in V_w} \hat{\mu} (e^{\ell_1}, e^{\ell_2}) d(e^{\ell_1} + b, e^{\ell_2})
\end{align*}
\end{small}
\paragraph{Combining Spatial and Temporal Dynamic Graphs.}

By adding a temporal dimension to our spatial dynamic graph, the resulting graph captures semantic changes of a mutual word translation pair $e_w^{\ell_1}$ and $e_w^{\ell_2}$ in languages $\ell_1$ and $\ell_2$ over time from $t-1$ to $t$---see Figure \ref{fig:tree-structure-spatial-temporal} (appendix).

To detect and compare semantic changes in $e_w^{\ell_1}$ and $e_w^{\ell_2}$, we undertake a two-fold process:
For each language, we employ a similarity matrix across sense clusters to detect the acquisition and loss of meanings in $e_w^{\ell_1}$ and $e_w^{\ell_2}$ over time, and then compare the detected changes along two dimensions:
\begin{itemize}
    \item  
    Consider that $e_w^{\ell_1}$ gains a new meaning $p_i^{\ell_1,t}$ at time $t$, $e_w^{\ell_2}$ another $p_j^{\ell_2,t}$. If 
    the semantic similarity, given by $s(p_i^{\ell_1,t}, p_j^{\ell_2,t})$, is greater than a cross-lingual threshold $t_{cs}$, then $e_w^{\ell_1}$ and $e_w^{\ell_2}$ are said to gain a new and similar meaning over time, thereby undergoing consistent acquisition changes in languages $\ell_1$ and $\ell_2$. Otherwise, their meaning changes over time diverge across languages.
    
    \item Consider that $e_w^{\ell_1}$ at time $t$ loses an existing meaning $p_i^{\ell_1, t-1}$, $e_w^{\ell_2}$ another $p_j^{\ell_2,t-1}$. If $s(p_i^{\ell_1, t-1}, p_j^{\ell_2,t-1}) > t_{cs}$, then $e_w^{\ell_1}$ and $e_w^{\ell_2}$ lose a similar meaning over time, thus undergoing consistent loss changes. Otherwise, their meaning changes differ across languages.
\end{itemize}

\begin{table*}[htbp]
\centering 
\footnotesize
\begin{tabular}{lccccc}
\toprule
    Approaches   & Avg & EN & DE & LA & SV \\
    \midrule
    \multicolumn{6}{l}{Static Word Embeddings}\\
    \midrule
    UWB \cite{prazak-etal-2020-uwb} & .687 & .622  &  .750  & \textbf{.700} &  .677 \\
    Life-Language \cite{asgari-etal-2020-emblexchange}  & .686 & .703 & .750 & .550 & .742 \\
    \midrule
    \multicolumn{6}{l}{Contextualized Word Embeddings}\\
    \midrule
    NLP@IDSIA \cite{kanjirangat-etal-2020-sst} & .637 & .622 & .625 &.625 &.677 \\   
    Skurt \cite{gyllensten2020sensecluster} & .629 & .568 & .562 & .675 & .710 \\
    Our Approach & \textbf{.776} & \textbf{.784} & \textbf{.813} & \textbf{.700} & \textbf{.806} \\
    \bottomrule
\end{tabular}
\caption{Accuracies from our approach and its counterparts in the SemEval2020 binary classification task.
}\label{tab:semantic-change-binary-results}
\end{table*}

\subsection{Our Clustering Method} 
\label{sec:clustering}
For each word, our goal is to partition a set of its contextualized word embeddings into multiple sense clusters.
To do this, we experimented with the popular $k$-means method widely adopted in previous works to produce sense clusters; however, we found that this method often produces poor sense clusters that fail to capture word senses, particularly problematic when dealing with low-frequency word senses (see Figure \ref{fig:kmeans-ours}). 
To address this, we present a clustering method that initializes each embedding as a separate cluster. We then iteratively merge two clusters whose centroids are of a distance smaller than a threshold\footnote{This threshold is tuned on the development sets that we created using ChatGPT. See \S\ref{app:lsc} for more details.} until no further pairs of such similar clusters can be found. The distance between a pair of cluster centroids $p_i$ and $p_j$ is given by $d(p_i, p_j)=\frac{1}{|C_w(p_i)| \cdot |C_w(p_j)|}\sum_{c_i\in C_w(p_i)}\sum_{c_j \in C_w(p_j)} d(c_i, c_j)$.
Our method allows for embeddings associated with different word senses (incl. low-frequency senses) to form their own sense clusters. 
To ensure quality, we exclude clusters with sizes below a threshold, which we consider as noisy clusters. 
Our method's procedure is provided in Algorithm \ref{alg:clustering}.

In our setup, we iterate through this procedure twice, applying different thresholds $t_{sc}$ and $t_{low}$ each time to achieve specific goals. In the first iteration, we generate a relatively large number of sense clusters for each word. This increases the chance for embeddings with low-frequency word senses to form their own clusters. We then detect and exclude unreliable low-frequency word sense clusters---which we consider as noisy clusters.
In the second iteration, we merge non-noisy clusters into only a few to capture word senses of each polysemous word. 

\begin{algorithm}[ht]
\centering
\footnotesize
\caption{Our Clustering Method}\label{alg:clustering}
\begin{algorithmic}[1]
\REQUIRE $\mathcal{C}_w=\{c_i\}_{i=1}^n$ as a set of contextualized embeddings of each word $w$, 
      $t_{sc}$ as the maximum distance between similar clusters,
      $t_{low}$ as the minimum cluster size for a low-frequency sense cluster.    
\STATE Initial centroids of clusters: $\mathcal{P}_w=\{p_i|p_i = c_i\}_{i=1}^n$
\WHILE{$\min_{p_i\in \mathcal{P}_w, p_j\in \mathcal{P}_w, i \neq j}d(p_i, p_j)  < t_{sc}$} 
\STATE $\mathcal{P}_w = \mathcal{P}_w \setminus \{p_i, p_j\} \cup \{\frac{p_i+p_j}{2}\}$ 
\ENDWHILE
\FOR{$p_i \in \mathcal{P}_w$}
\IF{$|\mathcal{C}_w(p_i)|  < t_{low}$}
\STATE $\mathcal{P}_w = \mathcal{P}_w \setminus \{p_i\}$
\ENDIF
\ENDFOR
\RETURN $\mathcal{P}_w$
\end{algorithmic}
\end{algorithm}

\section{Experiments}
\label{sec:experiments}
We evaluate our approach in SemEval2020 Task 1 (\S\ref{sec:lsc}), and showcase its ability as a visualization tool to detect semantic changes in both intra-language and inter-language setups (\S\ref{sec:explore}).
We provide analyses regarding our clustering method and embedding spaces---see \S\ref{app:analyses} (appendix).

\subsection{Intra-language Semantic Change}
\label{sec:lsc}
\paragraph{Setup.} 
We comparably evaluate our approach in SemEval2020 Task 1 for Unsupervised Lexical Semantic Change Detection \cite{schlechtweg2020semeval}. The task aims to detect intra-language semantic change over time through analyses across two corpora from the 19th and 20th centuries. The task encompasses two subtasks: binary classification and ranking across four languages, i.e., English (EN), German (DE), Latin (LA) and Swedish (SV).
We provide data statistics, task descriptions, our implementations details and selection of hyperparameters in \S\ref{app:lsc} (appendix). We use the last layer of m-BERT encoder \cite{devlin2018bert} to produce contextualized embeddings of target words across languages on the lemmatized corpora. 

\paragraph{Results.} 

Table \ref{tab:semantic-change-binary-results} compares our approach with its counterparts that rely on static and contextualized word embeddings in the SemEval2020 binary classification task (See \S\ref{app:ranking} for the results in the ranking task). We find that UWB and Life-Language based on static word embeddings outperform NLP@IDSIA and Skurt relying on contextualized embeddings. This unexpected result has been observed previously in \citet{schlechtweg2020semeval}, where the work attributes the underperformance of NLP@IDSIA and Skurt to the fact that they do not sufficiently leverage the power of contextualized embeddings. However, our approach based on contextualized embeddings largely outperforms all others, demonstrating its superiority in leveraging contextualized embeddings. The sources of our improvement are manifold: First, our approach includes a parameterized threshold that we use to stop our clustering process. This threshold is adjusted on the development sets that we created using ChatGPT, while previous approaches lack access to the development sets. Undoubtedly, our approach gains advantages from that, but more importantly, we argue that our improvement results from the careful design of our components---which we demonstrate through an ablation study.

\begin{figure*}[htbp]
\centering
\resizebox{0.8\textwidth}{!}{%
\begin{minipage}{0.3\textwidth}  
	\centerline{
        \subfigure[Time-dependent at time $t-1$]{
            \includegraphics[width=\linewidth]{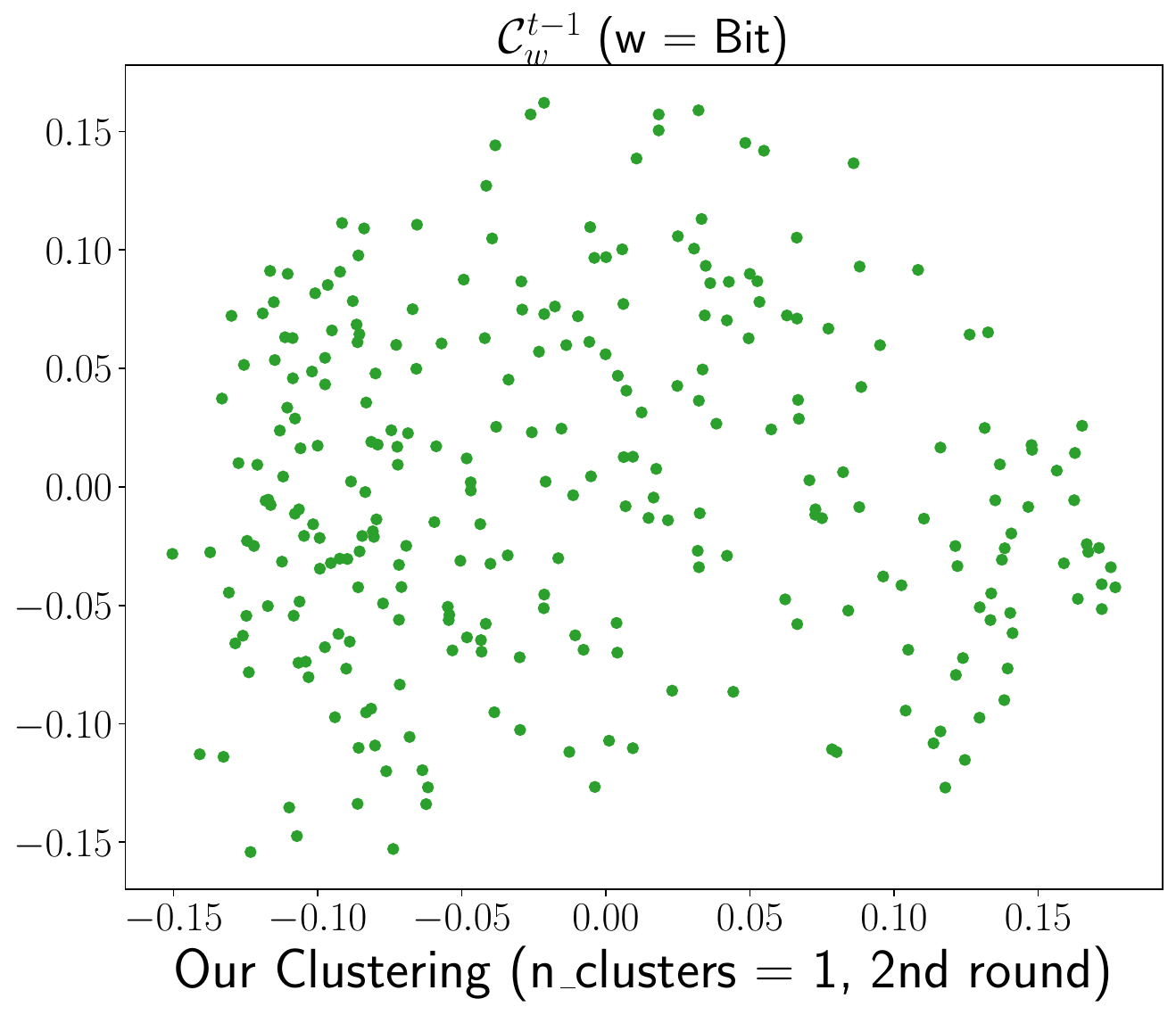}
        }
    } 
\end{minipage} 
\begin{minipage}{0.3\textwidth}  
	\centerline{
        \subfigure[Time-dependent at time $t$]{
            \includegraphics[width=\linewidth]{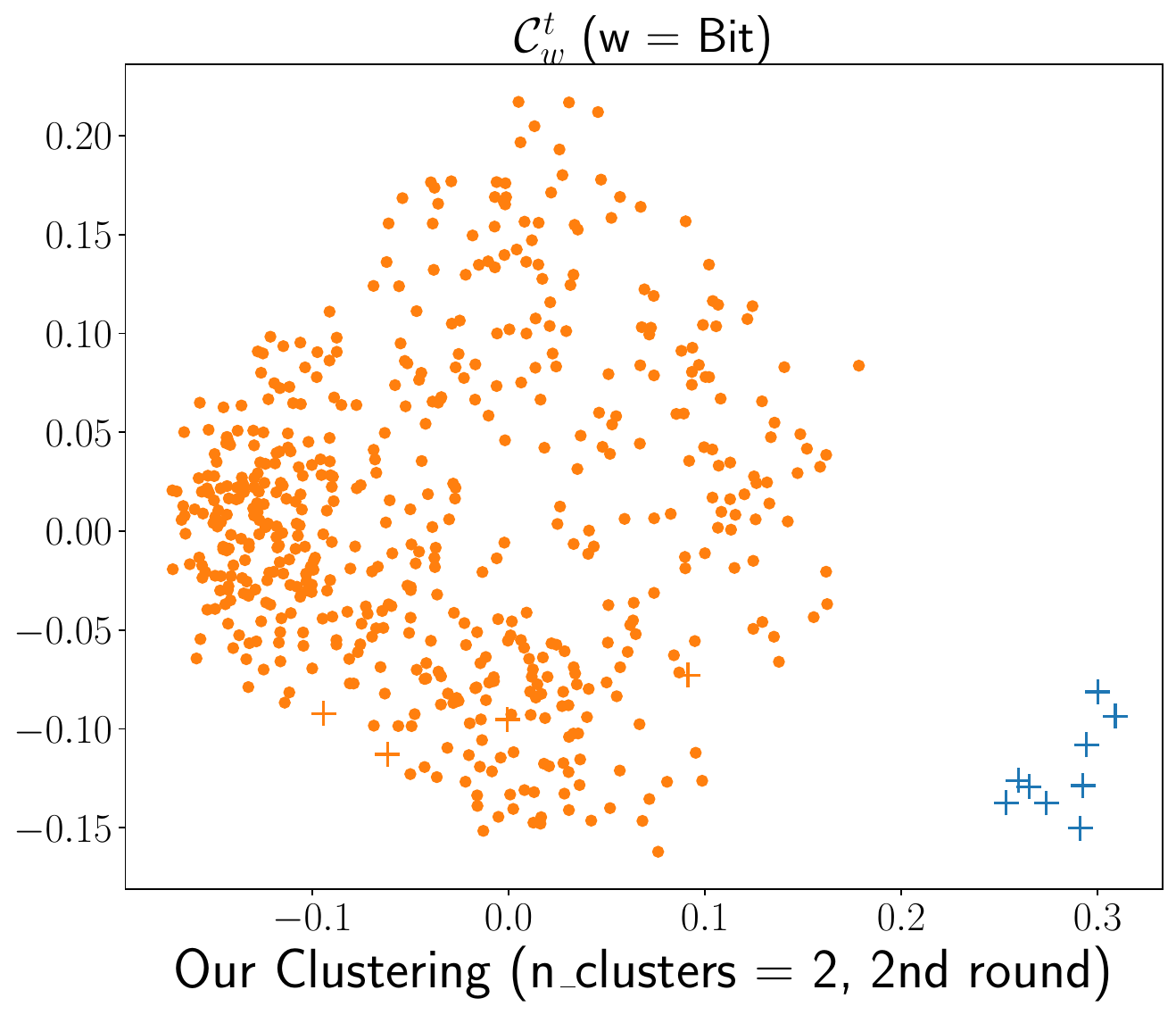}
        }
    } 
\end{minipage} 
\begin{minipage}{0.3\textwidth}  
	\centerline{
        \subfigure[Time-independent]{
            \includegraphics[width=\linewidth]{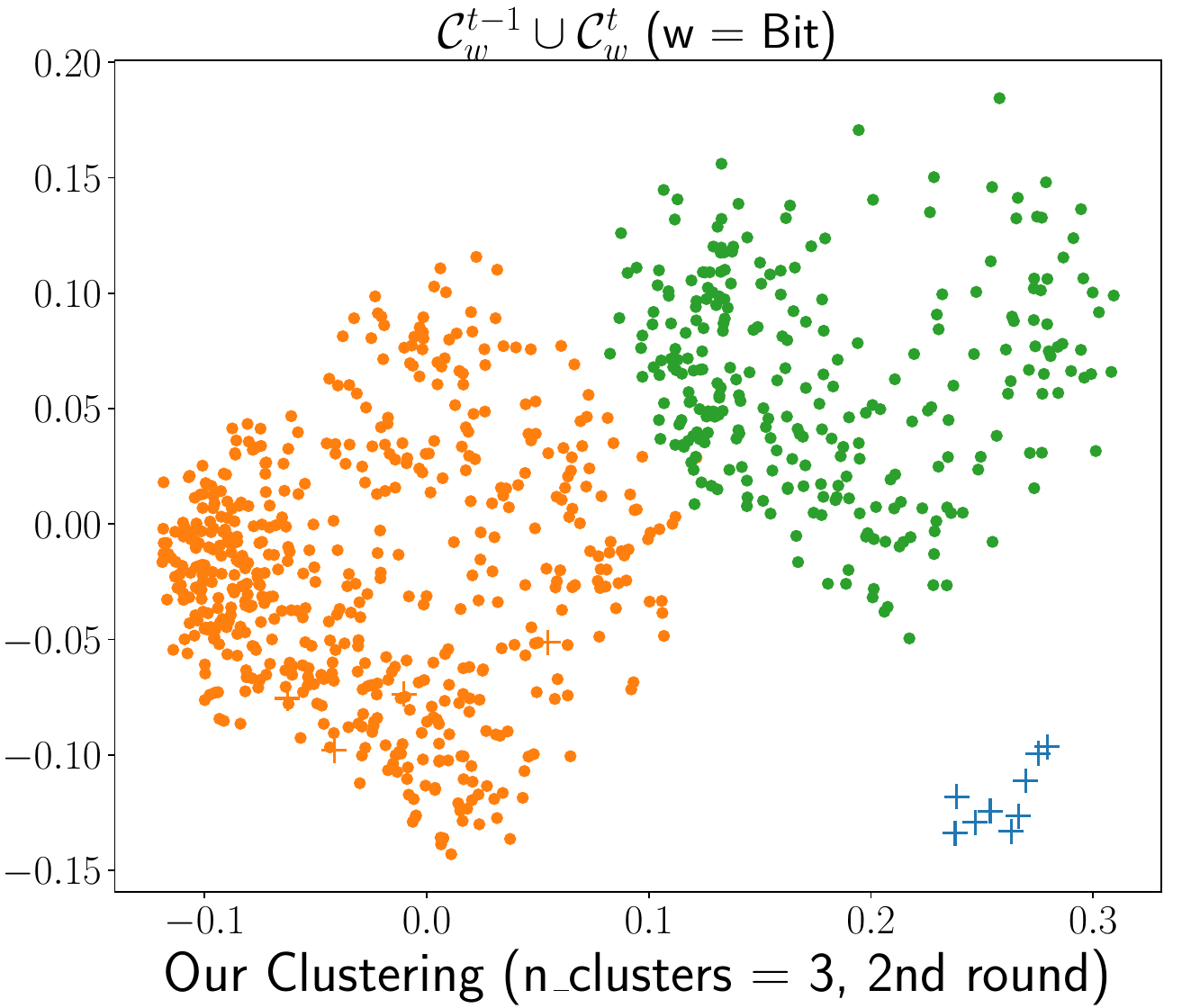}
        }
    }  
\end{minipage} 
}
\caption{Comparison of sense clusters for the word `bit' between the time-dependent (at time $t-1$ and $t$) and time-independent setups. Color indicates the cluster assignment of each point. A dot point represents the high-frequency word sense (a small piece), while a `+' indicates the low-frequency sense (binary digit).
}\label{fig:time-dep-indep}
\end{figure*}

\paragraph{Ablation Study.}  
Table \ref{tab:ablation} reports the ablation results on the three crucial components of our approach\footnote{We contrast the components of our approach with baseline approaches in Table \ref{tab:pipeline-comparison} (appendix).}.
\textbf{First}, we find that generating time-dependent sense clusters yields much better results than time-independent counterparts adopted in previous works \cite{kanjirangat-etal-2020-sst, gyllensten2020sensecluster}. We believe the previous approaches are based on the assumption that if a target word remains a consistent meaning over time, its word embeddings from different time periods should be grouped into a single time-independent sense cluster. However, this is challenging due to big context variations between the 19th and 20th corpora, causing contextualized encoders like BERT to misinterpret the consistent meaning as multiple dissimilar senses. As such, time-independent clusters misinterpret these senses over time as spurious meaning changes.
To support this hypothesis, we compare our clustering method between the time-independent and time-dependent setups. We see that Figure \ref{fig:time-dep-indep} (a)+(b), which produce sense clusters at each time period, adeptly capture the word senses of `bit' at each time. However, in Figure \ref{fig:time-dep-indep} (c) there are three distinct sense clusters, and the one in green, which has the same meaning of the orange one, is misinterpreted as a spurious new meaning by contextualized encoders. 

\begin{figure*}[htbp]
\centering
\resizebox{0.8\textwidth}{!}{%
\begin{minipage}{0.3\textwidth}  
	\centerline{
        \subfigure[Time-dependent at time $t$]{
            \includegraphics[width=\linewidth]{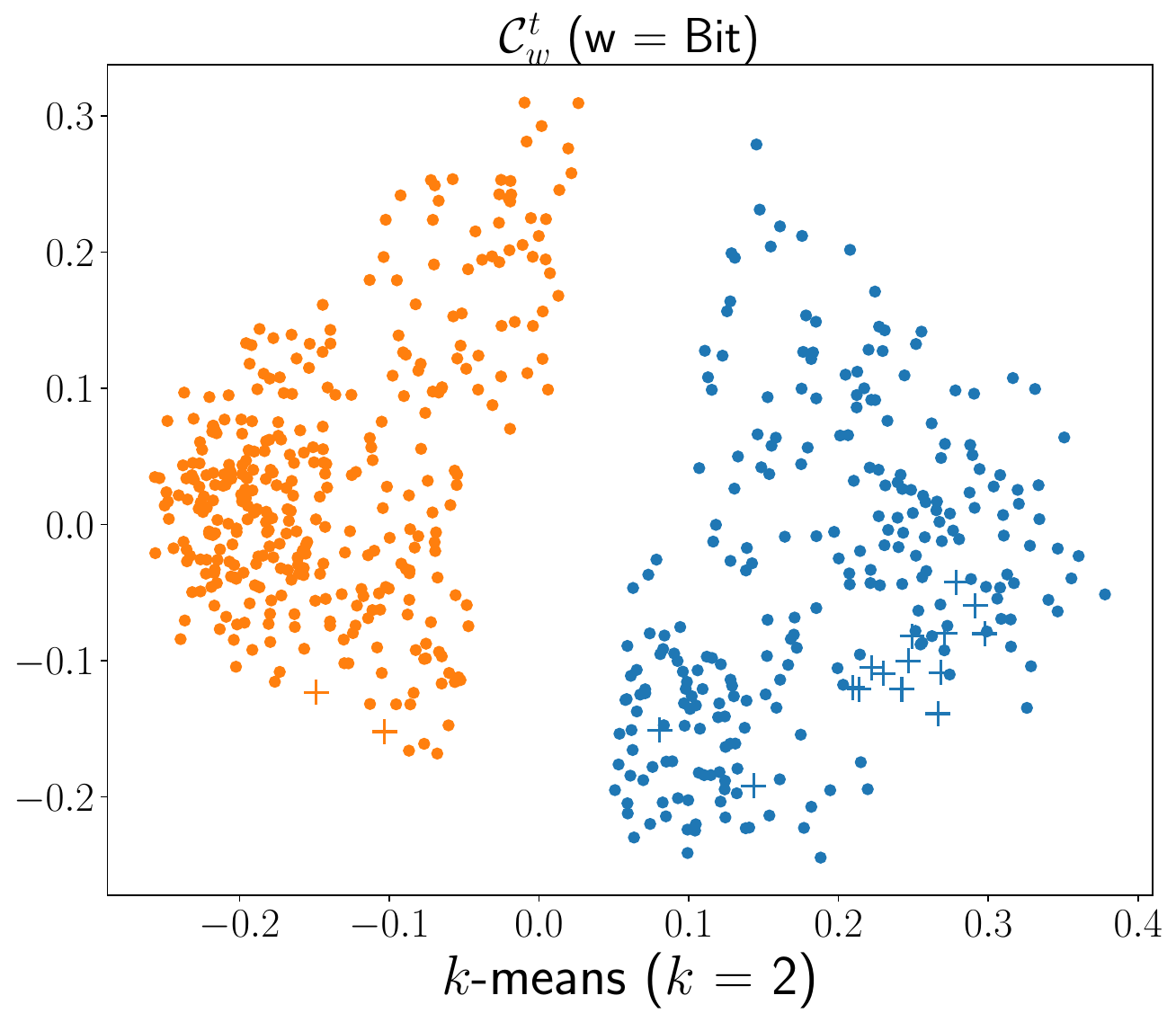}
        }
    } 
\end{minipage} 
\begin{minipage}{0.3\textwidth}  
	\centerline{
        \subfigure[Time-dependent at time $t$]{
            \includegraphics[width=\linewidth]{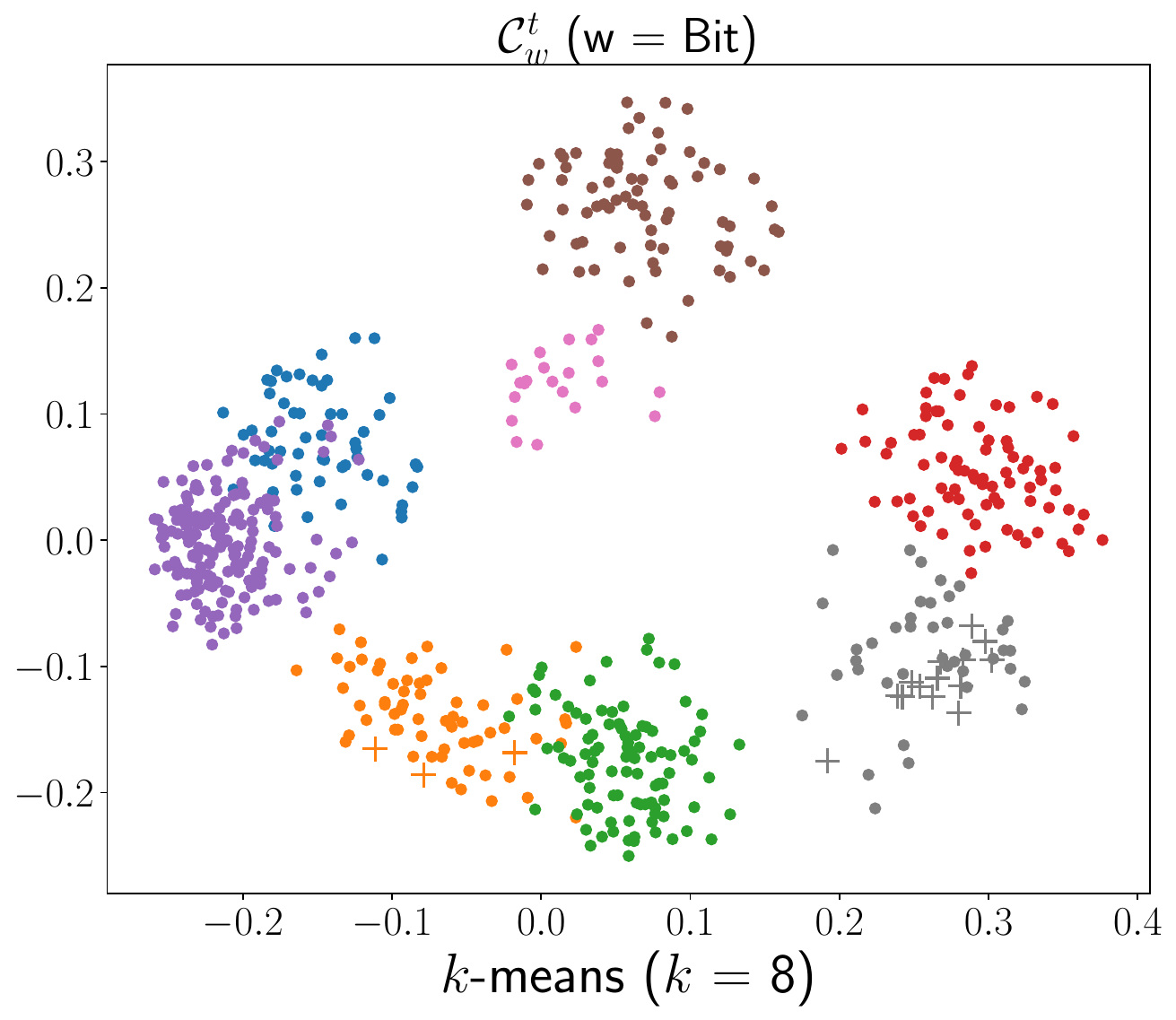}
        }
    } 
\end{minipage} 
\begin{minipage}{0.3\textwidth}  
	\centerline{
        \subfigure[Time-dependent at time $t$]{
            \includegraphics[width=\linewidth]{./pics/lowf-bit/bit_2nd_round_t2.pdf}
        }
    }  
\end{minipage} 
}
\caption{Comparison of sense clusters produced by $k$-means and our method. 
}\label{fig:kmeans-ours}
\end{figure*}

\textbf{Second}, we see that our clustering method considerably outperforms the popular $k$-means. The reasons for this are depicted in Figure \ref{fig:kmeans-ours}: (a) with $k=2$ (too small), most embeddings representing the low-frequency word sense (marked with `+') are wrongly subsumed into the high-frequency sense cluster in blue, and moreover, the two sense clusters in orange and blue share the same meaning and should not be separated; (b) with $k=8$ (too large): the high-frequency sense is wrongly divided into multiple sense clusters, despite low-frequency sense being correctly identified and mostly forming a distinct cluster in gray; (c) our clustering method produce two sense clusters that effectively capture both high-frequency and low-frequency senses. This is because our approach does not fix the number of clusters but instead leverage the idea of iteratively merging clusters until convergence, subject to some conditions. This provides the flexibility to find an adaptable number of clusters. 

\textbf{Lastly}, we see that neighbor-based distance metric greatly surpasses Euclidean distance. Unlike Euclidean distance, which quantifies the similarity between sense clusters by computing the similarity between cluster centroids, our neighbor-based metric does this by computing the similarity between the $k$ semantically nearest neighbors to each cluster centroid. 
We believe that our metric, which leverages $k$ neighbors rather than just the centroid, allows us to better capture the semantics of sense clusters, providing a more accurate reflection of the similarity between sense clusters.

\begin{table}
\footnotesize
\centering
\setlength{\tabcolsep}{5pt}
\begin{tabular}{@{}llc@{}}
\toprule
Components &  Approaches & EN \\
\toprule
All-in-one & Our Approach & .784 \\
Clustering Strategy & $\ominus$ Time-dep.  $\oplus$ Time-indep. & .649 \\
Clustering Method & $\ominus$ Our method  $\oplus$ $k$-means & .649 \\
Distance Metric & $\ominus$ Neighbor-based $\oplus$ Euclidean & .676 \\
\bottomrule
\end{tabular}
\caption{Ablation test in the SemEval2020 binary classification task, where $\ominus$ X $\oplus$ Y means the replacement of component X in our approach by component Y.}\label{tab:ablation}
\end{table}

\subsection{Exploratory Study}
\label{sec:explore}
Our setup is detailed in \S\ref{app:exploratory} (appendix).

\paragraph{Intra-Language Semantic Change.} 
Consider the polysemous English word `mouse'. It is well-known that the word's meaning has evolved from a small rodent to a computer input device over time.
Figure \ref{fig:temporal-graph} showcases the ability of our temporal dynamic graph, which is derived from our approach on both historical and present English corpora, to capture the recorded semantic changes of the word `mouse' over time. We find that the word initially has only one meaning represented by its five-nearest neighbors such as `rat' and `bat' at time $t-1$. As time progresses, the word maintains this original meaning while gaining a new meaning about computer device at time $t$---characterized by its corresponding neighbors in blue color. We find many such examples across languages, and provide a few in Figure \ref{fig:sc-intralingual-en-appendix} (appendix).
\label{sec:inter-lang}

\begin{figure*}[htbp]
\centering
\resizebox{0.9\textwidth}{!}{%
\begin{minipage}{0.33\textwidth}  
	\centerline{\includegraphics[width=\linewidth]{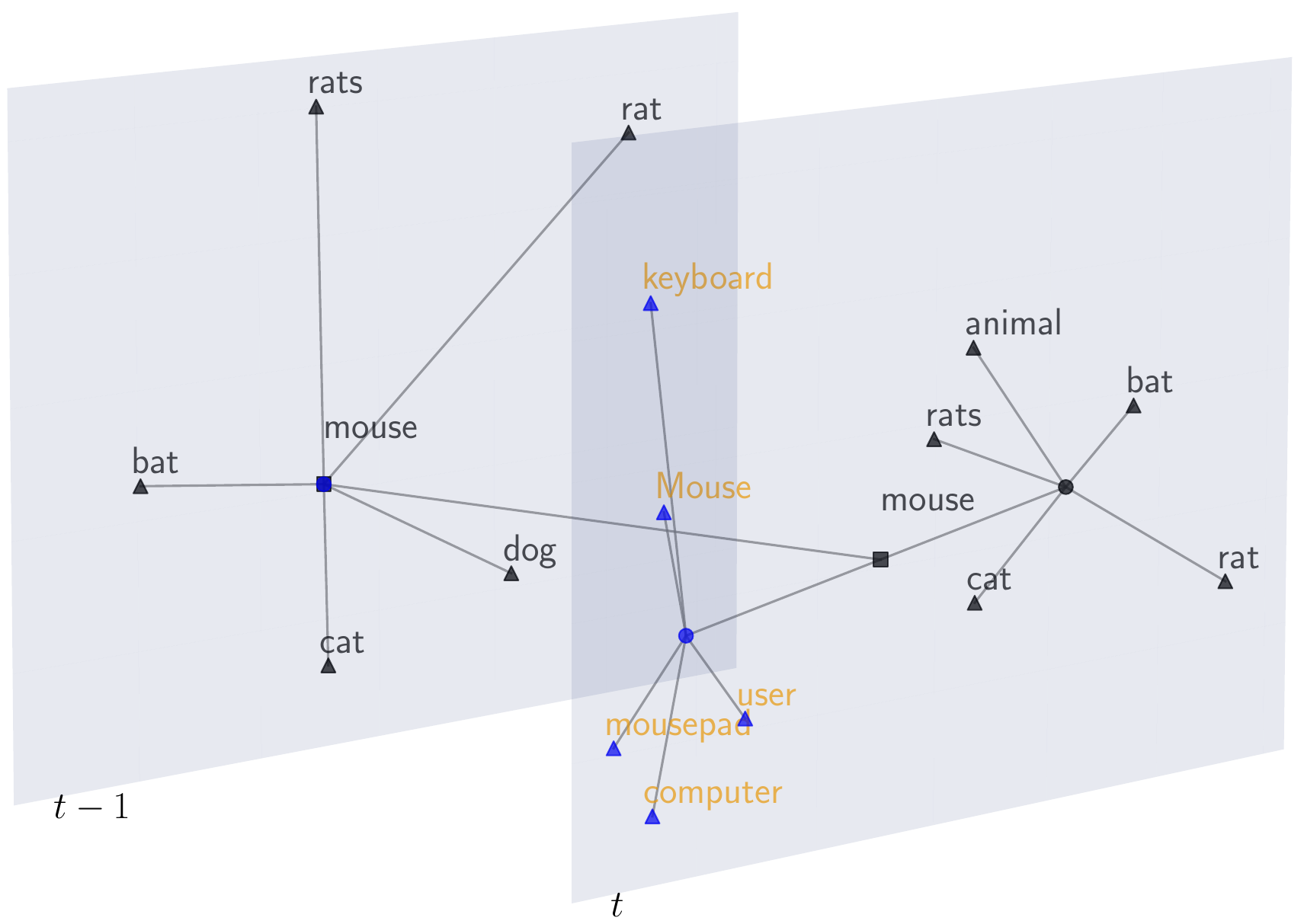}} 
\end{minipage} 
\begin{minipage}{0.33\textwidth}  
	\centerline{\includegraphics[width=\linewidth]{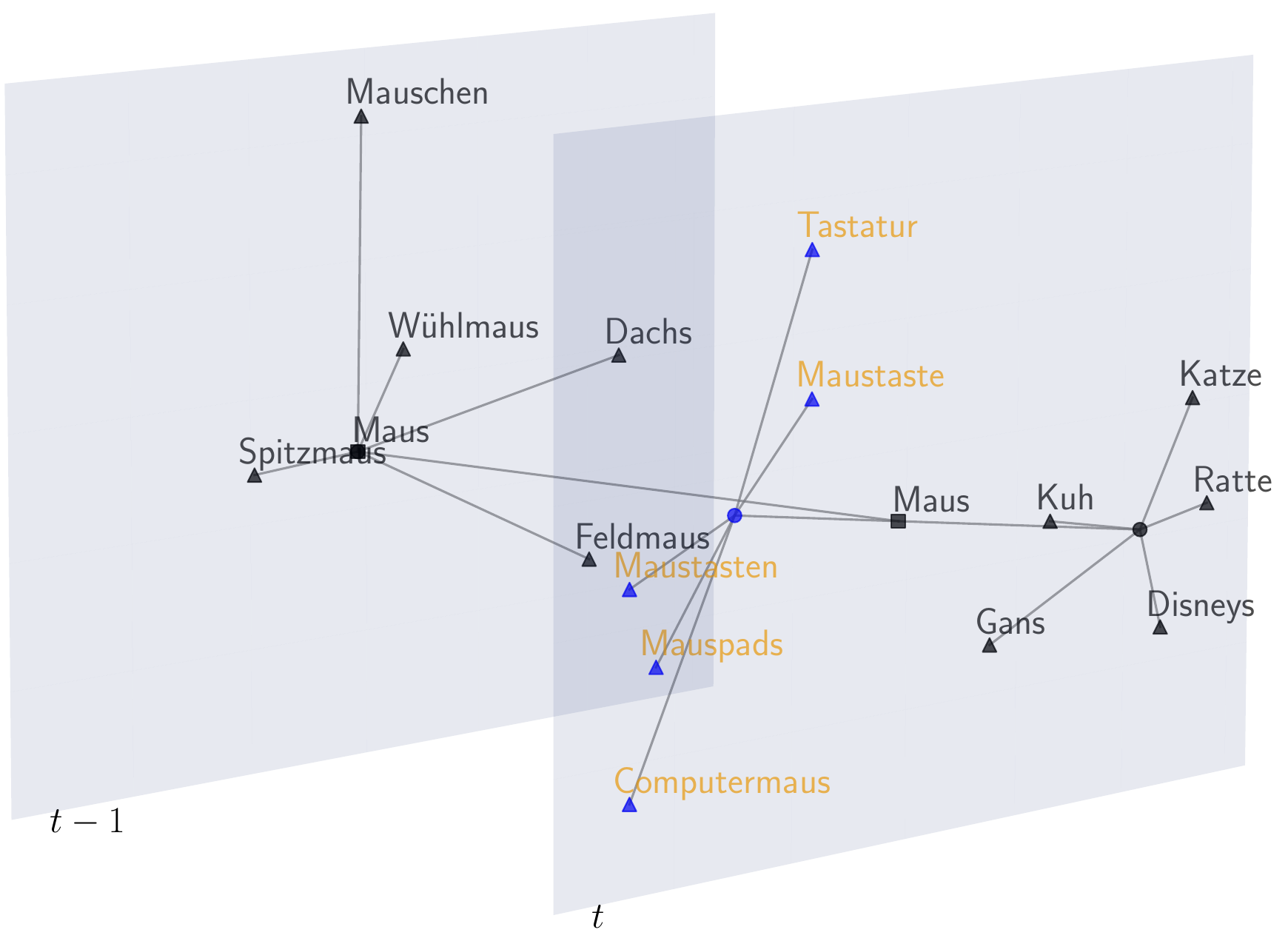}}  
\end{minipage} 
\begin{minipage}{0.33\textwidth}  
	\centerline{\includegraphics[width=\linewidth]{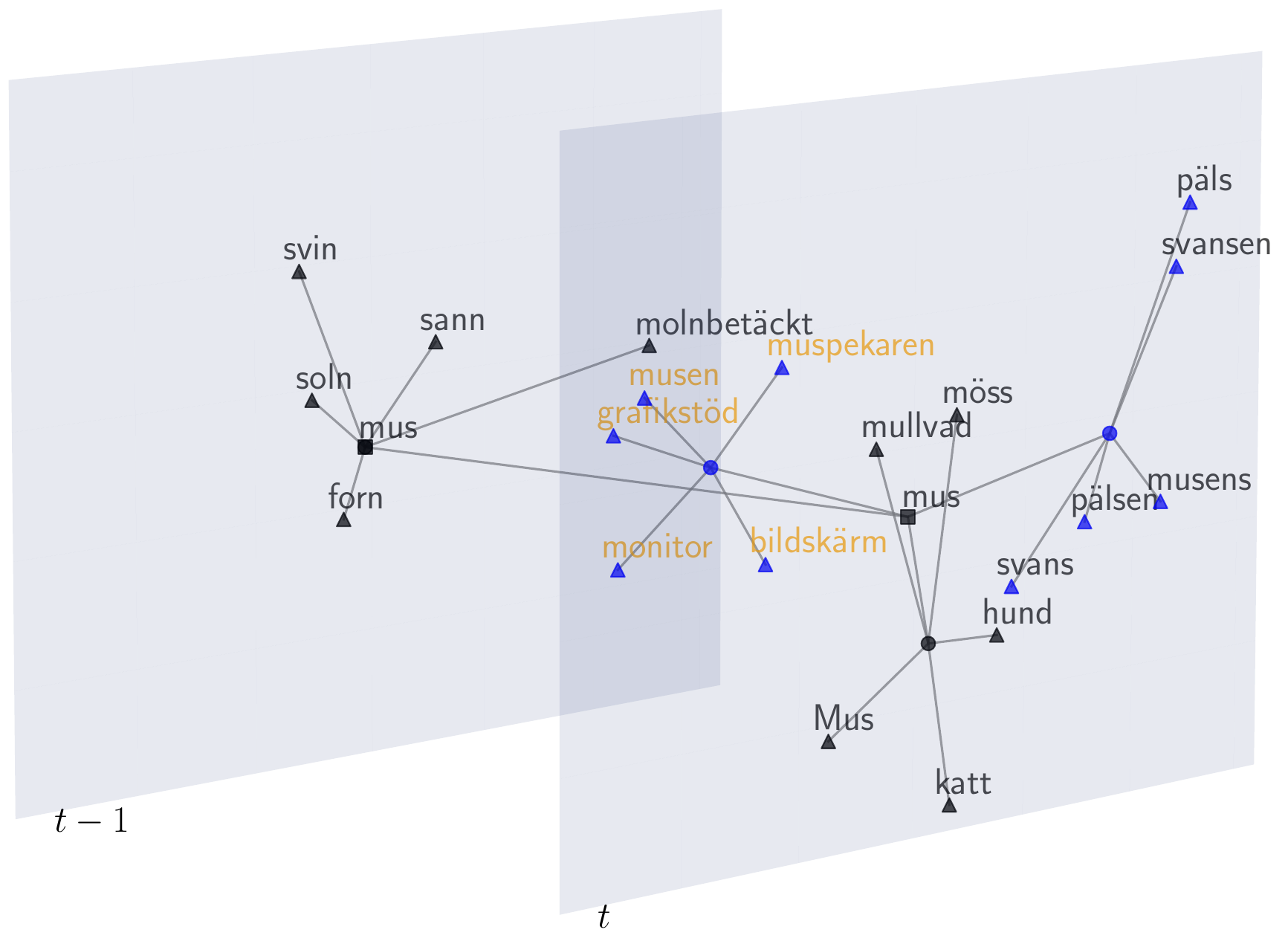}}  
\end{minipage} %
}
\caption{Representation of the detected semantic changes for the word translations 'mouse', 'Maus', and 'mus' in three temporal dynamic graphs. Blue nodes indicate the acquisition of new meanings, while orange text additionally marks certain new meanings that are considered highly similar, i.e., undergoing consistent acquisition change.}\label{fig:sc-xlingual}
\end{figure*}

\paragraph{Inter-Language Semantic Changes.} 
Figure \ref{fig:sc-xlingual} compares the detected semantic changes for the word translations 'mouse', 'Maus', and 'mus' across English, German and Swedish over time. We observe that each of these word translations holds just a single meaning at time $t-1$, within the 19-century historical corpus. However, at time $t$ within the current Wikipedia corpus, `mouse' and `Maus' gain new meanings, indicated in blue nodes, while `mus' gains two new meanings, similarly indicated. Furthermore, we note that the blue nodes with the same meaning of ``computer device'' are labeled in orange text across languages: Both `mouse' and `Maus' acquire the same new meaning, implying that these two words undergo consistent acquisition changes over time. However, the meaning changes of these two words diverge from that of `mus': Although all three words acquire the same meaning ``computer device'', `mus` gains another new meaning related to `svans` and `päls' at time $t$.
This example showcases the potential of our approach as a visualization tool 
to detect semantic divergence and consistency across languages over time. We provide examples regarding meaning loss in Figure \ref{fig:sc-xlingual-appendix} (appendix). 

We validated these results on both Wiktionary and the etymological dictionary\footnote{\url{https://www.etymonline.com/}}. While the overall results are accurate, the neighbors connected to each root node do not necessarily represent synonyms of that node. For instance, in Figure 7 (left), both `cat' and `dog' are closer than `rat' to `mouse'. This only means that the contexts in which `cat' and `dog' appear are more similar to the context of `mouse'.

\section{Conclusions}

We proposed a graph-based clustering approach to capture changes within each word sense across both time and languages.
We addressed an intriguing concern that contextualized embeddings coupled with clustering methods seem not suitable for detecting semantic change, as they underperform their static embedding counterparts.
We identified a crucial reason for this: Previous approaches rely on low-quality clustering methods to handle contextualized embeddings. 
Our results demonstrated that, when equipped with an appropriate clustering method, strategy, and distance metric, contextualized embeddings can produce high-quality sense clusters that effectively capture word senses, even low-frequency ones. These factors attribute to our approach's superiority over the shared task winner, ULB, which relies on static embedding. Further, the use of our approach as a visualization tool highlights its value in conducting exploratory studies on both intra- and inter-language semantic changes. 

\section{Limitations}
Our approach still lags behind static embedding counterparts in the ranking task for languages other than English (see \S\ref{app:ranking}). Further improvements may result from improving the quality of embeddings for non-English languages. 

Lack of a standard evaluation setup poses a challenge in tracking recent progress in the \textbf{intra-language} setup.
For instance, 
\citet{card-2023-substitution} reported results in SemEval2020 and GEM ranking tasks; \citet{teodorescu2022ualberta} did in LSCDiscovery binary and ranking tasks; we did in SemEval2020 classification and ranking tasks.

Further, the absence of benchmark datasets for detecting the divergence of semantic changes in the \textbf{inter-language} setup poses another challenge in evaluating our approach. Moreover, in the cross-lingual setup, we addressed the misalignment between the embedding spaces of two languages; however, our focus was on word-level rather than  meaning-level alignments. Thus, it remains unclear how the embedding spaces (adjusted via word-level alignments) handle words with polysemy profiles. These present avenues for future work.

\section{Ethical Considerations}
Our work proposed an approach based on BERT to detect semantic change and evaluated the approach on the historical datasets from SemEval2020 Task 1. We acknowledge the potential biases arising from both our approach and the datasets. In historical corpora, a bias towards male authors is often observed. Regarding our approach, BERT is known to encode social biases related to gender and race. Up to now, it remains unclear how these biases may affect the results of semantic change detection. We leave this question to future work.

\section*{Acknowledgements}
We thank the anonymous reviewers for their thoughtful comments that greatly improved the texts. We also thank Dominik Schlechtweg and Steffen Eger for providing feedback on the early version of this work.
This work has been supported by the Klaus Tschira Foundation and Young Marsilius Fellowship,
Heidelberg, Germany. 

\bibliography{custom}
\bibliographystyle{acl_natbib}

\clearpage
\appendix

\section{Appendix}
\label{sec:appendix}

\subsection{Ranking Task.}
\label{app:ranking}
\paragraph{Our approach.} We describe our approach used to perform the SemEval2020 ranking task. Following many works \cite{schlechtweg2020semeval,kanjirangat-etal-2020-sst,kutuzov2020uio}, we design a criterion to grade the degree of semantic change by comparing the frequencies of word meanings across time periods. Such a criterion allows for capturing both past and prospective changes in word meanings. For instance, when comparing the frequencies of a word meaning over time, a frequency decline from time $t-1$ to $t$ suggests the potential loss of the meaning in the future. Here, we aim to measure the degree of both meaning acquisition and loss over time. Our criterion is detailed below:

Once the sets of cluster centroids $\mathcal{P}_{w}^{t-1}$ and $\mathcal{P}_{w}^{t}$ at time $t-1$ and $t$ are determined\footnote{In contrast to the binary classification setup, our clustering method does not exclude outliers in the ranking setup.}, we then divide the combined set of these centroids into $m$ clusters. Each cluster $H_i$ represents a distinct word sense over time, and comprises centroids from different time periods that exhibit high similarities exceeding a threshold $t_{sc}$. This implies that the associated meanings of these centroids remain unchanged over time. For each target word, we let $A^t=(a_1^t, \dots, a_m^t)$ denote the frequency distribution of $m$ word senses at time $t$, where $a_i^t = \frac{\sum_{p \in H_i \cap \mathcal{P}_{w}^{t}} |C_w^t(p)|}{\sum_{p\in \mathcal{P}_{w}^{t}}|C_w^t(p)|}$, and similarity for $A^{t-1}$. By comparing the two frequency distributions, we illustrate three scenarios:

\begin{itemize}
    \item $a_i^{t-1} = 0$ and $a_i^{t} > 0$: acquiring a new meaning at time $t$.
    \item $a_i^{t} = 0$ and $a_i^{t-1} > 0$: losing an existing meaning at $t-1$.
    \item $a_i^{t} > 0 $ and $a_i^{t-1} > 0$: indicating the degree of meaning change over time.
\end{itemize}

Follow \citet{schlechtweg2020semeval}, we grade the degree of semantic change by computing the Jensen-Shannon distance between two frequency distributions, noted as $\mathrm{JSD} (A^t, A^{t-1})$ in our setup. 

\paragraph{Results.}
Table \ref{tab:semantic-change-rank-results} compares the results of our approach and its counterparts in the SemEval2020 ranking task.
We see that our approach performs best among the approaches relying on contextualized word embeddings. Our approach substantially outperforms the recent substitution-based approach in 3 out of 4 languages, and surpasses static embedding counterparts in English. However, our approach still lags behind in other languages; interestingly, it outperforms static embedding counterparts in all languages for binary classification. Our analysis on this is the following: 

First, the ranking task is inherently more challenging, as it requires to quantify the fine-grained degree of semantic change. Second, m-BERT is known to produce different embedding quality across languages, with superior embedding quality in English. In binary classification, where the task is straightforward, embedding quality matters little. However, for the challenging ranking task, lower-quality embeddings can harm the results. We leave the verification of this hypothesis to future work.

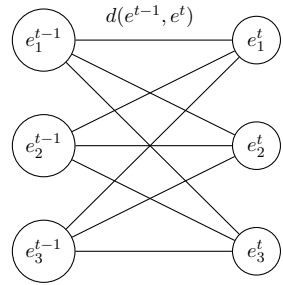
\begin{figure}
    \centering
\begin{tikzpicture}[scale=0.7, transform shape,]
  \foreach \i in {1,2,3} {
    \node[circle, draw] (L\i) at (0, -2*\i) {$e_\i^{t-1}$};
  }
  \foreach \i in {1,2,3} {
    \node[circle, draw] (R\i) at (4,- 2*\i) {$e_\i^{t}$};
  }
  \draw[-] (L1) -- (R1);
  \draw[-] (L1) -- (R2);
  \draw[-] (L1) -- (R3);
  \draw[-] (L2) -- (R1);
  \draw[-] (L2) -- (R2);
  \draw[-] (L2) -- (R3);
  \draw[-] (L3) -- (R1);
  \draw[-] (L3) -- (R2);
  \draw[-] (L3) -- (R3);
  \node[] at (2, -1.55) {$d(e^{t-1}, e^{t})$};
\end{tikzpicture}
\caption{Illustration of bipartite matching for computing the similarity between sense clusters' centroids $p_1^t$ and $p_1^{t-1}$. Here, $\{e_i^{t}\}_{i=1}^3$ indicates the representative embeddings of three semantically nearest neighboring words to $p_1^t$. The same applies to $\{e_i^{t-1}\}_{i=1}^3$ and $p_1^{t-1}$.}
\label{fig:matching}
\end{figure} 

\begin{table*}
\centering 
\footnotesize
\begin{tabular}{lccccc}
\toprule
    Approaches   & Avg & EN & DE & LA & SV \\
    \midrule
    \multicolumn{6}{l}{Static Word Embeddings}\\
    \midrule
    UG\_Student\_Intern \cite{pomsl2020circe} & \textbf{.527} & .422 & \textbf{.725} & .412 & .547 \\
    Jiaxin \& Jinan \cite{zhou2020temporalteller}  & .518 & .325 & .717 & .440 & \textbf{.588}\\
    \midrule
    \multicolumn{6}{l}{Contextualized Word Embeddings}\\
    \midrule
    Substitution \cite{card-2023-substitution} & .488 & .547 & .563 & \textbf{.533} & .310 \\   
    Skurt \cite{gyllensten2020sensecluster} & .374 & .209 & .656 & .399 & .234 \\
    Our Approach & .506 & \textbf{.569} & .656 & .377 & .423 \\
    \bottomrule
\end{tabular}
\caption{Results from our approach and its counterparts in the SemEval2020 ranking task. Results are reported in Spearman correlation. }\label{tab:semantic-change-rank-results}
\end{table*}

\begin{table*}[h]
\centering 
\footnotesize
\setlength\tabcolsep{3pt} 
\begin{tabular}{lccccccccccr}
\toprule
\multicolumn{1}{l}{} & \multicolumn{5}{c}{Corpus \#1}                & \multicolumn{5}{c}{Corpus \#2}                    & \multicolumn{1}{l}{} \\
 \cmidrule(lr){2-6}\cmidrule(lr){7-11}
Language             & period ($t-1$)    & \#tokens & avg/t & max/t & min/t & period ($t$)        & \#tokens & avg/t & max/t & min/t & \#targets              \\
 \cmidrule(lr){1-1} \cmidrule(lr){2-6}\cmidrule(lr){7-11} \cmidrule(lr){12-12}
English              & 1810-1860 & 25,955  & 701   & 4,211  & 86    & 1960-2010     & 30,060  & 812   & 4,062  & 106   & 37                   \\
German               & 1800-1900 & 71,556  & 1,490  & 28,756 & 35    & 1946-1990     & 42,260  & 880   & 8,539  & 103   & 48                   \\
Latin                & 200BC-1BC & 27,548  & 688   & 3,498  & 26    & 100AD-present & 129,568 & 3,239  & 10,362 & 245   & 40                   \\
Swedish              & 1790-1830 & 35,021  & 1,129  & 6,934  & 83    & 1895-1903     & 126,126 & 4,068  & 14,583 & 89    & 31 \\          \bottomrule
\end{tabular}
\caption{Statistics of the SemEval2020 Task 1 Corpus.  The last column `\#targets' denotes the number of target words, while the column `\#tokens' denotes the total token count of target words. The column `avg/t' indicates the average token count of each target word, while the column `max/t' indicates the maximum token count per target word, and the column `min/t' indicates the minimum token count per target word. }\label{tab:semeval-dataset-statistic}
\end{table*}

\subsection{Experimental Setups for SemEval2020 Task 1}
\label{app:lsc}

\paragraph{Datasets.} Table \ref{tab:semeval-dataset-statistic} provides data statistics for the SemEval2020 Task 1.

\paragraph{Task Descriptions.} SemEval2020 Task 1 consists of two subtasks, namely (a) binary classification, where one decides whether the meaning of each target word has changed over time by analyzing word usage across two text corpora from different time periods and (b) ranking, where a list of provided target words should be ranked based on scores given by a criterion indicating the degree to which each word undergoes semantic change.

\paragraph{Implementation Details.} For each language, we produce contextualized word embeddings of target words from two time periods of text corpora, and then employ our clustering method to partition embeddings of each target word into multiple sense clusters in order to constitute a temporal dynamic graph. As mentioned previously, we iterate through our clustering procedure twice. This requires two chosen hyperparameters in each iteration: (a) $t^0_{low}$ and $t^1_{low}$, representing the minimum occurrence for a low-frequency meaning, i.e., the minimum cluster size and (b) $t^0_{sc}$ and $t^1_{sc}$, representing the maximum distance between similar clusters. Furthermore, we need to leverage bipartite matching based on $k$-nearest semantic neighbors (with $k$ as an extra hyperparameter) to compute the similarity between sense clusters. This step is crucial for detecting nuanced meaning changes over time.  

We use a grid search to tune the following two hyperparameters on \textbf{the development set we constructed using ChatGPT} in each language: $t^0_{sc} \in \{t| 0.10 < t < 0.35, 100\cdot t \in N_{+} \}$ and $t^1_{sc} \in \{t| 0.10 < t < 0.45, 100\cdot t \in N_{+} \}$. In all setups, we set $t^0_{low}$ to 5 and 
consider clusters with sizes below 5 as noisy clusters\footnote{In the SemEval 2020 shared task, a new word sense is acknowledged upon meeting two rules: (a) this sense associates with fewer than 2 word tokens at time t-1 and (b) it associates with more than 5 word tokens at time t. If a word sense meets (a) but violates (b), for example, having less than 5 word tokens at time t, then this new sense is considered unacceptable and categorized as a noisy sense. We follow this idea and remove sense clusters with word tokens fewer than 5.}. We set $t^1_{low}$ to 0, as noisy clusters should have been removed when the first iteration ends. We set $k$ to 14---see our clustering analysis in \S\ref{app:analyses}.
Our configuration of these hyperparameters across languages are reported in Table \ref{tab:config-classification} 
and \ref{tab:config-ranking} for classification and ranking tasks.
We note that the chosen $t^1_{sc}$ is applied to our detection criterion for finding similar and dissimilar sense clusters, i.e., to detect the presence of semantic change in each word sense.

\begin{table}[ht]
\centering 
\footnotesize
\setlength\tabcolsep{5pt} 
\begin{tabular}{lccccc}
    \toprule
    Languages    & $t^0_{sc}$ &  $t^1_{sc}$ & $k$ & $t^0_{low}$ &  $t^1_{low}$\\
    \midrule
    English & 0.34 & 0.40  & 14  & 5 & 0 \\
    German  & 0.22 & 0.38  & 14  & 5 & 0  \\
    Latin  & 0.16 & 0.16  & 14  & 5 & 0\\
    Swedish  & 0.28 & 0.32  & 14  & 5 & 0\\
    \bottomrule
\end{tabular}
\caption{Configuration of hyperparameters across languages in the SemEval2020 binary classification task.}\label{tab:config-classification}
\end{table}

\begin{table}[ht]
\centering 
\footnotesize
\setlength\tabcolsep{5pt} 
\begin{tabular}{lccccc}
    \toprule
    Languages    & $t^0_{sc}$ &  $t^1_{sc}$ & $k$ & $t^0_{low}$ &  $t^1_{low}$\\
    \midrule
    English & 0.34 & 0.40  & 14  & 0 & 0 \\
    German  & 0.22 & 0.38  & 14  & 0 & 0  \\
    Latin  & 0.16 & 0.16  & 14  & 0 & 0\\
    Swedish  & 0.28 & 0.32  & 14  & 0 & 0\\
    \bottomrule
\end{tabular}
\caption{Configuration of hyperparameters across languages in the SemEval2020 ranking task.}\label{tab:config-ranking}
\end{table}

\begin{table*}[]
\centering 
\footnotesize
\setlength\tabcolsep{3pt} 
\begin{tabular}{lccccccccccr}
\toprule
\multicolumn{1}{l}{} & \multicolumn{5}{c}{Corpus \#1}                & \multicolumn{5}{c}{Corpus \#2}                    & \multicolumn{1}{l}{} \\
 \cmidrule(lr){2-6}\cmidrule(lr){7-11}
Language             & period ($t-1$)    & \#tokens & avg/t & max/t & min/t & period ($t$)        & \#tokens & avg/t & max/t & min/t & \#targets              \\
 \cmidrule(lr){1-1} \cmidrule(lr){2-6}\cmidrule(lr){7-11} \cmidrule(lr){12-12}
English              & 1810-1860 & 3,294  & 329   & 947 & 23    & Wiki (08.2023)     & 53,019  & 5,301   & 23,733  & 755   & 10                   \\
German               & 1800-1900 & 17,893 & 1,789  & 4,536 & 84    & Wiki (08.2023)    & 43,963  & 4,396   & 22,809  & 240   & 10                   \\
Swedish              & 1790-1830 & 12,409  & 1,240  & 5,310  & 14    & Wiki (08.2023)     & 47,629 & 4,762  & 35,249 & 25    & 10 \\          \bottomrule
\end{tabular}
\caption{Statistics of the corpus in the inter-language setup. }\label{tab:inter-dataset-statistic}
\end{table*}

\paragraph{Construction of Development Sets using ChatGPT.}
As SemEval2020 Task 1 operates in an unsupervised
setting, the task lacks development sets, with the
entire corpus treated as evaluation sets. Here, we create a development set per language on which we tune the hyperparameters of our clustering approach for each language. Each development set includes 8 target words that are unseen in evaluation sets. Each target word associates with two senses. We use ChatGPT-3.5 to produce 100 sentences that contextualize each sense. We now describe our data construction approach in detail:

For each target word, we begin by instructing ChatGPT to provide a list of possible word senses, and then verify their accuracy using Wiktionary. After that, we select two verified word senses from the list and instruct ChatGPT to generate a corpus of $n$ sentence that evenly incorporate both word senses of the target word. Then, we construct a gold label vector, denoted as $\mathcal{Y}_w = [y_1, \dots, y_n]$ with $y_i \in \{0,1\}$, where $y_i$ specifies whether the target word in the $i$-th sentence within the corpus corresponds to the first or second word sense.

We observed that ChatGPT yields sentences of satisfactory quality, which contains expected word meanings of each target word and requires only minor human corrections such as the need for extra instructions to generate longer sentences. As an example, Table \ref{tab:gpt-gen-corpus} reports our instructions for the word ``ratio'' with a specific sense in Latin. 

\begin{table*}
\centering 
\footnotesize
\setlength\tabcolsep{5pt} 
\renewcommand\arraystretch{1.3} 
\begin{tabular}{ll}
\toprule
Role    & Conversations                                                                                                                                                                         \\
\midrule
Human    & Please create sentences for the Latin word `ratio' that has the meaning of `reason or rationality, the ability to think \\
        & logically and make sound judgments' in the format: a Latin sentence enclosed in double quotation marks, followed \\ 
        & by an English translation enclosed in parentheses.                      \\
ChatGPT & 1. "Sua ratione et prudentia, ille vir difficiles decisiones sapienter adhibebat."                                                                                                \\
        & (By his reason and prudence, that man wisely made difficult decisions.)                                                                                                          \\
        & 2. "Ratione uti poteris, neque impulsu emotionum agere debes."                                                                                                                    \\
        & (You should use reason, not be driven by the impulse of emotions.)                                                                                                               \\
        & ... ...(sentences examples)                                                                                                                                                                              \\
\hdashline
Human    & Good. Please create a bit longer sentences.                                                                                                                                             \\
ChatGPT & 1. "Cum homines ratione utantur, societates humanae pacem et progressum promovere possunt."                                                                                       \\
        & (When people use reason, human societies can promote peace and progress.)                                                                                                        \\
        & 2. "Ratione, quam discimus, in omni aspectu vitae uti debemus, ut optima decisiones capiamus."                                                                                    \\
        & (We should use the reason we learn in every aspect of life to make the best decisions.)                                                                                          \\
        & ... ...(sentences examples)                                                                                                                                                                            \\
\hdashline
Human    & Good. Please write more sentences in the same format.                                                                                                                                                        \\
ChatGPT & ... ...(sentences examples)                                                                                                                                                                \\

\bottomrule
\end{tabular}
\caption{Our instructions for creating sentences that contain the Latin word `ratio' with a specific word sense. 
}\label{tab:gpt-gen-corpus}

\end{table*}

Recall that our clustering approach involves two hyperparameters $t^0_{sc}$ and $t^1_{sc}$ to determine whether two clusters are similar enough to be merged. To tune these hyperparameters on the development sets we constructed, we first use our clustering approach to produce a prediction of label vector, denoted as $\mathcal{\hat{Y}}_w(t^0_{sc}, t^1_{sc}) = [\hat{y}_1, \dots, \hat{y}_n]$ with $\hat{y}_i \in \{0,1\}$ for each configuration of hyperparameters. Then, we use grid search to tune the hyperparameters based on the idea of Adjusted Mutual Information (AMI) \cite{vinh2009information}, denoted as:
\begin{align*}
\argmax_{t^0_{sc} \in (0,1), t^1_{sc} \in (0,1)} \sum_{w \in W_{\ell}}  \mathrm{AMI}(\mathcal{Y}_w, \mathcal{\hat{Y}}_w(t^0_{sc}, t^1_{sc}))
\end{align*}
where $W_\ell$ denotes a set of target words for each language $\ell$.

\subsection{Experimental Setups for Exploratory Study}
\label{app:exploratory}

\paragraph{Datasets.}
We choose a set of target word triplets that are translations in English, German and Swedish, such as  \{`mouse', `Maus', `mus'\}. For each of these languages, we consider the SemEval2020 corpus specific to that language from the earlier time period (the 19 century) as the historical corpus at time $t-1$. For the present corpus at time $t$, we opt for a random selection of the most recent Wikipedia dump, rather than the SemEval2020 corpus from the later time period. This is because the later time periods in the three languages are substantially different, making it unreliable to compare semantic changes across languages. We provide data statistics in Table \ref{tab:inter-dataset-statistic}. 

\paragraph{Implementation Details.}
Regarding the choice of our hyperparameters, we apply the same cross-lingual threshold $t_{cs}$ to all languages, and set the threshold to the  threshold $t_{sc}$ used in the English intra-language setup, denoted as $t_{cs}^{\mathrm{All}} = t_{sc}^{\mathrm{EN}}$. The reason for this is the following: Since we re-align the embedding spaces of target languages (German and Swedish) to the source language English, we apply the hyperparameters that were tuned on the English development set to all languages in the inter-language setup. We set $k$ to 14 for $k$-nearest neighboring words. 

\begin{table}[ht] 
\centering 
\footnotesize
\setlength\tabcolsep{2pt} 
\begin{tabular}{cccccccc}
    \toprule
   ($\ell_1$, $\ell_2$)   & $m(e^{\ell_1}, \mathcal{C}_w^{\ell_1})$ & $m(e^{\ell_1}, \mathcal{C}_w^{\ell_2})$ & $m(e^{\ell_1}, \mathcal{C}_w^{\ell_2} + b)$ \\
    \midrule
    (EN, DE) & 0.64  & 0.46 & 0.64 \\
    (EN, SV) & 0.64  & 0.45  & 0.65  \\
    \bottomrule
\end{tabular}
\caption{Results of embedding space alignment. }\label{tab:mdiff}
\end{table}

\begin{table*}[ht]
\footnotesize
\centering
\setlength\tabcolsep{3pt}
\begin{tabular}{lll}
\toprule
 & \multicolumn{2}{c}{\textbf{Contextualized Word Embeddings}} \\
 \midrule
\textbf{Components} & \textbf{Previous Approaches} & \textbf{Our Approach} \\
\midrule
Clustering Method & $k$-means & our clustering method \\
\midrule
Clustering Strategy & time-independent sense clusters & time-dependent sense clusters \\
\midrule
Semantic Representation & word embeddings & graph \\
\midrule
Distance Metric & Euclidean distance & neighbor-based distance \\
\midrule
Detection Criterion & frequency-based criterion & similarity between sense clusters \\
\bottomrule
\end{tabular}
\caption{Contrasting previous approaches \cite{kanjirangat-etal-2020-sst, gyllensten2020sensecluster} and our approach for detecting semantic change. All these approaches combine contextualized word embeddings with a clustering method but differ in several aspects. Our neighbor-based metric is adopted in two components: our clustering method and detection criterion.}
\label{tab:pipeline-comparison}
\end{table*}

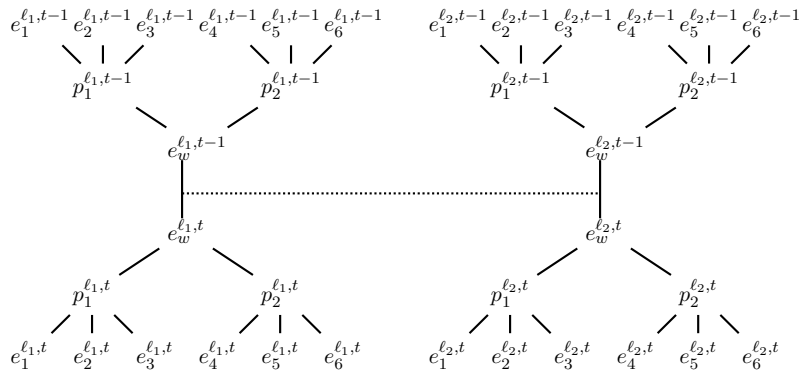
\begin{figure*}[ht]
    \centering
\begin{tikzpicture}
[thick,scale=0.55, every node/.style={scale=0.75}]
\node [right=4cm]  at (0,-1)  {$e_w^{\ell_1, t-1}$}[rotate=180]
  child {node {$p_2^{\ell_1, t-1}$}
    child {node {$e_6^{\ell_1, t-1}$}}
    child {node {$e_5^{\ell_1, t-1}$}}
    child {node {$e_4^{\ell_1, t-1}$}}
    }
  child [missing] {}  
  child [missing] {}  
  child {node {$p_1^{\ell_1, t-1}$}
    child {node {$e_3^{\ell_1, t-1}$}}
    child {node {$e_2^{\ell_1, t-1}$}}
    child {node {$e_1^{\ell_1, t-1}$}}
    };

\draw (6,-1.3) -- (6,-2.7);

\node[right=4cm] at (0,-3) {$e_w^{\ell_1, t}$}
  child {node {$p_1^{\ell_1, t}$}
    child {node {$e_1^{\ell_1, t}$}}
    child {node {$e_2^{\ell_1, t}$}}
    child {node {$e_3^{\ell_1, t}$}}
    }
  child [missing] {}  
  child [missing] {}  
  child {node {$p_2^{\ell_1, t}$}
    child {node {$e_4^{\ell_1, t}$}}
    child {node {$e_5^{\ell_1, t}$}}
    child {node {$e_6^{\ell_1, t}$}}
    };

\node [right=4cm]  at (10,-1)  {$e_w^{\ell_2, t-1}$}[rotate=180]
  child {node {$p_2^{\ell_2, t-1}$}
    child {node {$e_6^{\ell_2, t-1}$}}
    child {node {$e_5^{\ell_2, t-1}$}}
    child {node {$e_4^{\ell_2, t-1}$}}
    }
  child [missing] {}  
  child [missing] {}  
  child {node {$p_1^{\ell_2, t-1}$}
    child {node {$e_3^{\ell_2, t-1}$}}
    child {node {$e_2^{\ell_2, t-1}$}}
    child {node {$e_1^{\ell_2, t-1}$}}
    };

\draw (16,-1.3) -- (16,-2.7);

\draw[densely dotted] (6,-2.1) -- (16,-2.1);

\node[right=4cm] at (10,-3) {$e_w^{\ell_2, t}$}
  child {node {$p_1^{\ell_2, t}$}
    child {node {$e_1^{\ell_2, t}$}}
    child {node {$e_2^{\ell_2, t}$}}
    child {node {$e_3^{\ell_2, t}$}}
    }
  child [missing] {}  
  child [missing] {}  
  child {node {$p_2^{\ell_2, t}$}
    child {node {$e_4^{\ell_2, t}$}}
    child {node {$e_5^{\ell_2, t}$}}
    child {node {$e_6^{\ell_2, t}$}}
    };    
\end{tikzpicture}
    \caption{Representation of semantic changes of a mutual word translation pair over time in a temporal and spatial dynamic graph that links two temporal dynamic graphs in different languages.}
    \label{fig:tree-structure-spatial-temporal}
\end{figure*}

\subsection{Analyses}
\label{app:analyses}

\paragraph{Comparing Clustering Approaches.} We comparably evaluate two classes of clustering approaches: (a) explicit/predetermined choice of the number clusters: K-means and Gaussian Mixture\footnote{\url{https://scikit-learn.org/stable/modules/generated/sklearn.mixture.GaussianMixture.html}} and (b) implicit choice: Affinity Propagation \cite{frey2007clustering} and our approach. To begin, we select 8 target words in each language. Each word has two word senses with uneven frequency distribution (100:20). We use the popular metric Purity Scoring\footnote{\url{https://nlp.stanford.edu/IR-book/html/htmledition/evaluation-of-clustering-1.html}}) to evaluate clustering quality---the higher purity score indicates better quality. In Table \ref{tab:clustering_performance}, we see that our approach is quite advantageous in this setup, demonstrating its ability to capture both high and low-frequency word senses. In English, we see the performance gain of our approach is comparatively smaller. This might be because m-BERT is known to produce higher-quality embeddings in English compared to other languages, making it less susceptible to the poor quality of baseline clustering approaches.

\begin{table}[h]
\centering
\footnotesize
\begin{tabular}{lcccc}
\toprule
Algorithms & EN & DE & LA & SV \\ 
\toprule
K-means & 0.975 & 0.778 & 0.664 & 0.775 \\
Gaussian Mixture & 0.939 & 0.775 & 0.670 & 0.754 \\
Affinity Propagation & 0.891 & 0.741  & 0.686 & 0.662 \\
Our Clustering & \textbf{0.994} & \textbf{0.879} & \textbf{0.877} & \textbf{0.909} \\
\bottomrule
\end{tabular}
\caption{Purity scores across four approaches in the 100:20 frequency distribution setup.}
\label{tab:clustering_performance}
\end{table}

\begin{figure*}[htbp]
\centering
\subfigure[English]{
\includegraphics[width=5.5cm]{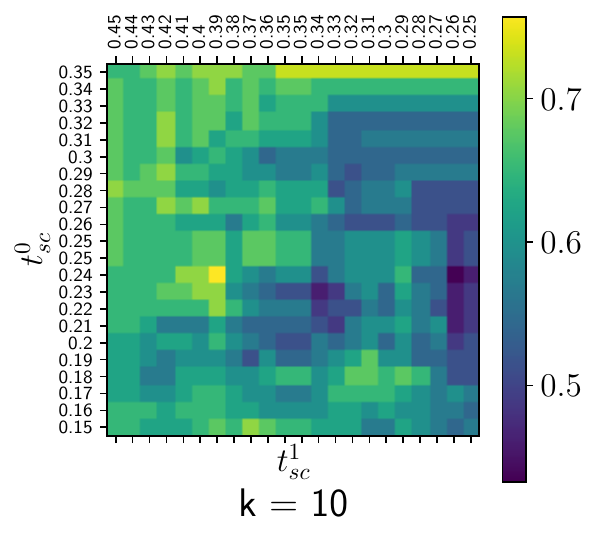}
\includegraphics[width=5.5cm]{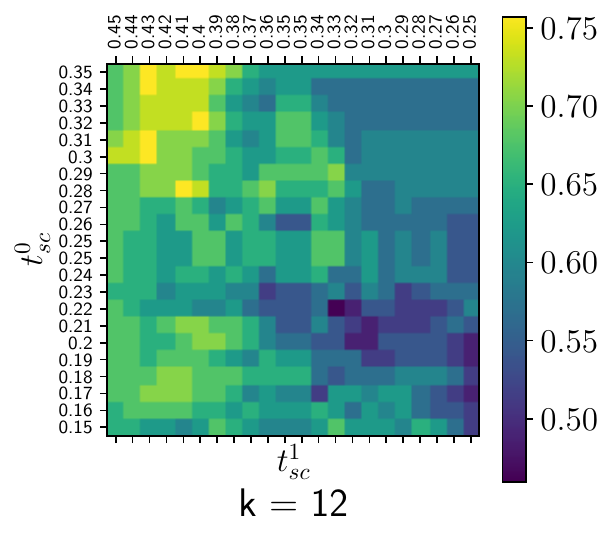}
\includegraphics[width=5.5cm]{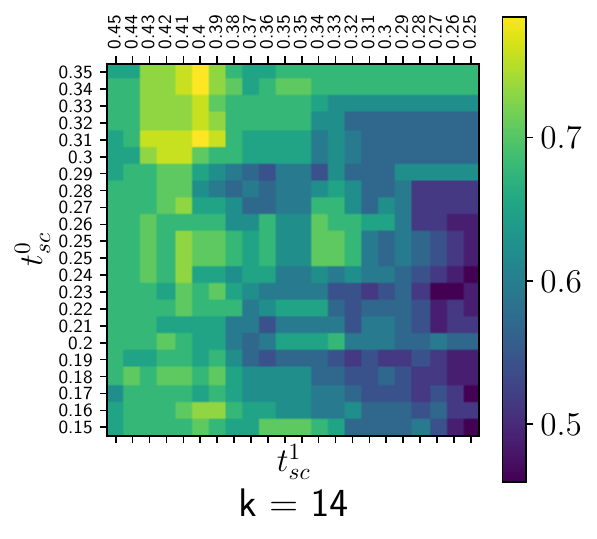}
}
\quad
\subfigure[German]{
\includegraphics[width=5.5cm]{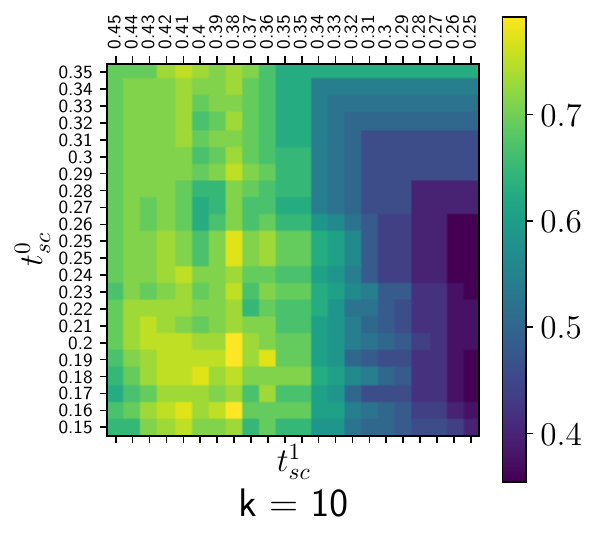}
\includegraphics[width=5.5cm]{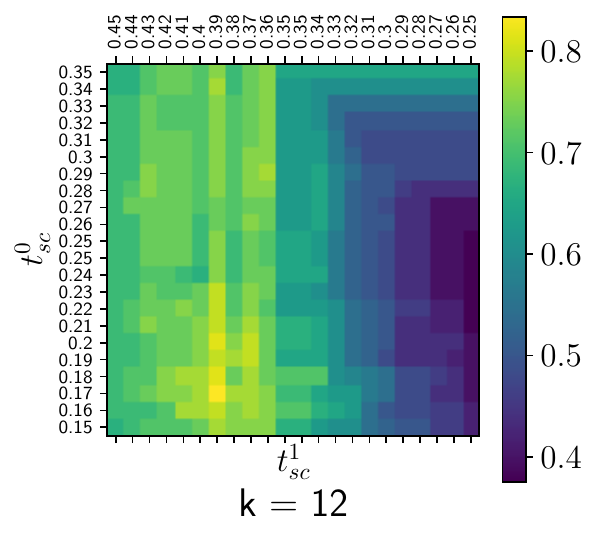}
\includegraphics[width=5.5cm]{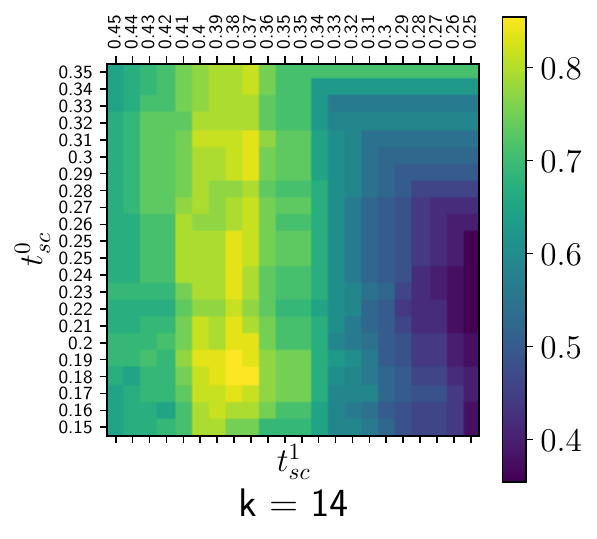}
}
\quad
\subfigure[Latin]{
\includegraphics[width=5.5cm]{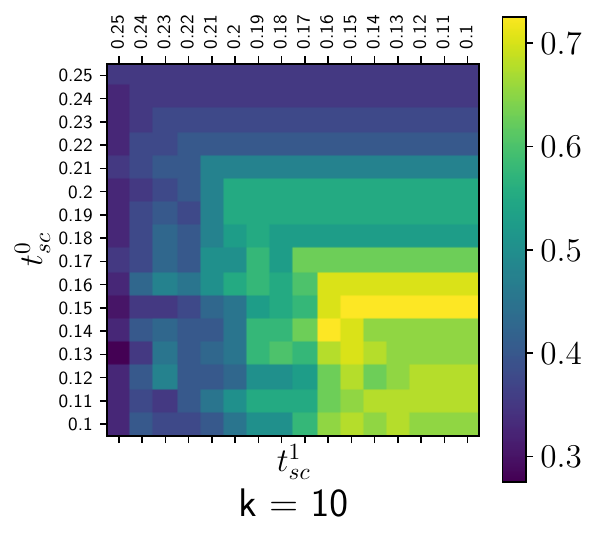}
\includegraphics[width=5.5cm]{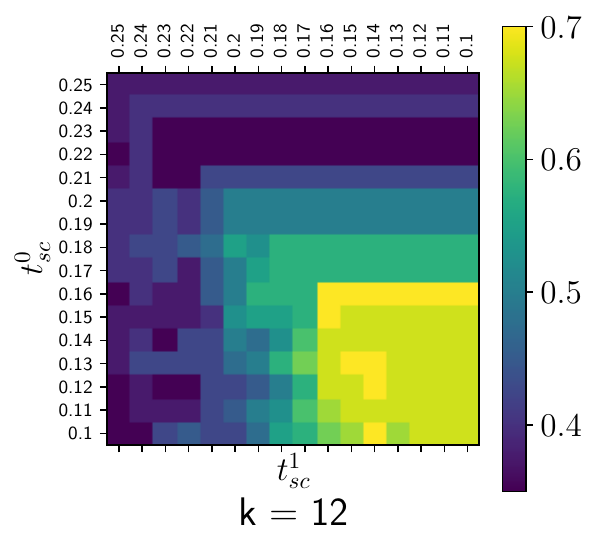}
\includegraphics[width=5.5cm]{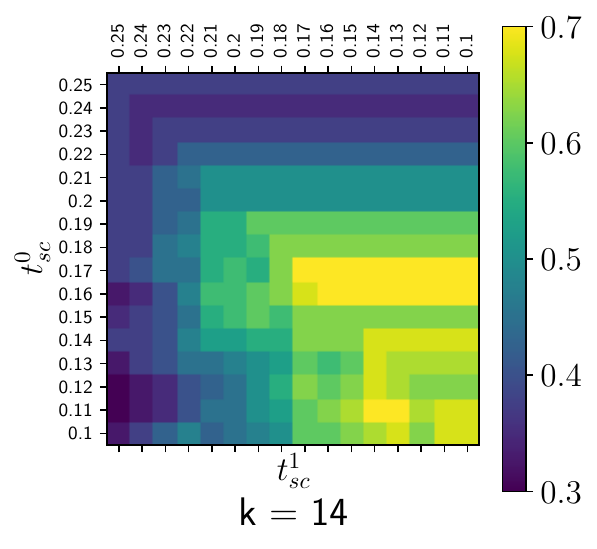}
}
\quad
\subfigure[Swedish]{
\includegraphics[width=5.5cm]{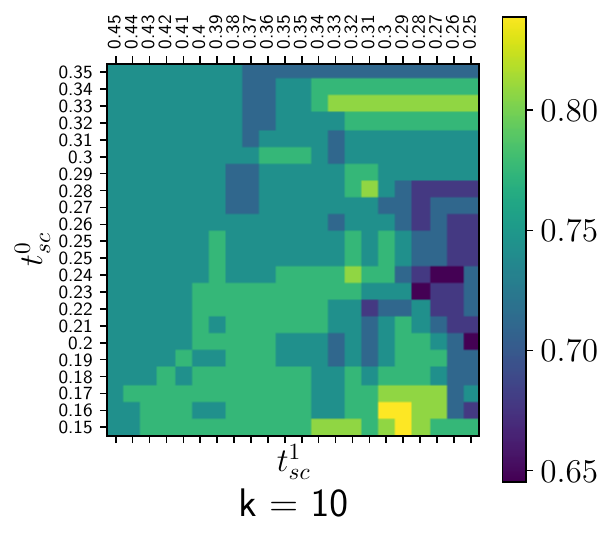}
\includegraphics[width=5.5cm]{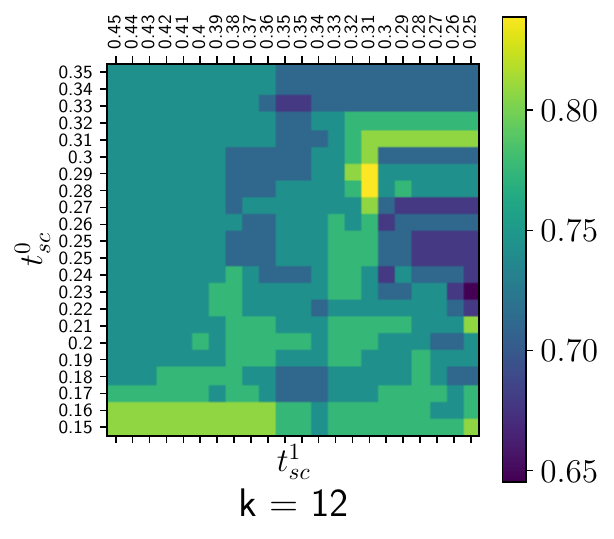}
\includegraphics[width=5.5cm]{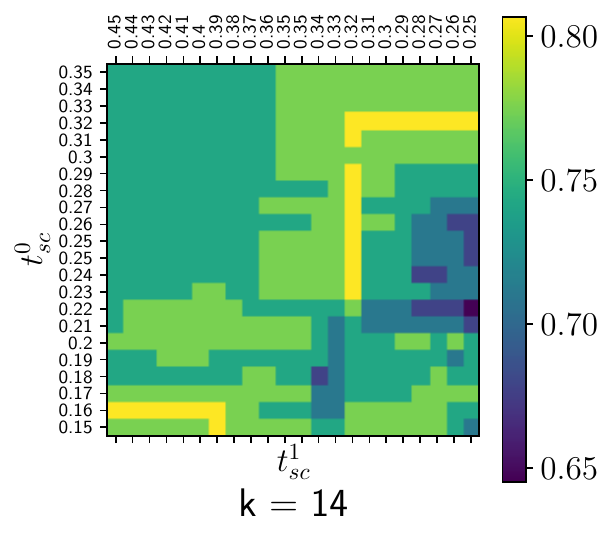}
}

\caption{Relationships between the threshold values and accuracies in the SemEval2020 binary classification task.}
\label{semeval-threshold-binary-acc}
\end{figure*}

\paragraph{Our Clustering Method.}
Figure \ref{semeval-threshold-binary-acc} shows the relationships between the choice of thresholds ($t^0_{sc}$ and $t^1_{sc}$) and the corresponding detection accuracy. We find that the high accuracy area colored in bright yellow expands greatly as $k$ increases, particularly for English and German. This means that the more nearest semantic neighboring words are involved, the higher detection accuracy our approach achieves. Furthermore, we see that the brightest areas across languages are shown in different locations, and these areas associate with very different configurations of $t^0_{sc}$ and $t^1_{sc}$, even for typologically similar language pairs such as English and German. This is because the SemEval2020 corpora in English and German are collected from different time periods (see Table \ref{tab:semeval-dataset-statistic}), making these two languages further apart from each other.

\paragraph{Bilingual Embedding Spaces.}

If two embedding spaces of languages $\ell_1$ and $\ell_2$ align well, they should share the same space centroid and the same topological structure. We consider the topological structure of the $\ell_1$ embedding space as $m(e^{\ell_1}, \mathcal{C}_w^{\ell_1})=\frac{1}{|\mathcal{C}_w^{\ell_1}|}\sum_{c_i \in \mathcal{C}_w^{\ell_1}} s(e^{\ell_1}, c_i)$, i.e, the average similarity between each point in $\mathcal{C}_w^{\ell_1}$ and the $\mathcal{C}_w^{\ell_1}$'s centroid $e^{\ell_1}$. Therefore, $m(e^{\ell_1}, \mathcal{C}_w^{\ell_1})$ should closely match $m(e^{\ell_1}, \mathcal{C}_w^{\ell_2})$ in this case.
However, Table \ref{tab:mdiff} shows that the score $m(e^{\ell_1}, \mathcal{C}_w^{\ell_2})$ is much lower than $m(e^{\ell_1}, \mathcal{C}_w^{\ell_1})$, implying that the ${\ell_1}$ and ${\ell_2}$ embedding spaces exhibit quite different topological structures. This arises from the fact that the two embedding spaces are initially misaligned. After applying a rectified vector $b$, we see a close match between $m(e^{\ell_1}, \mathcal{C}_w^{\ell_1})$ and $m(e^{\ell_1}, \mathcal{C}_w^{\ell_2} + b)$ in terms of topological structure, demonstrating the effectiveness of the chosen rectification approach \cite{liu2020study} for addressing  the misalignment between embedding spaces of different languages.

\subsection{Hardware Specifications and Execution Times}
All experiments were executed on a computer featuring an AMD CPU with 8 cores, 32GB of RAM and a single RTX3060 GPU with 12GB of memory. 
For each target word, it takes about 60 seconds for m-BERT to generate its contextualized word embeddings within 800 sentences on GPU; our clustering method takes about 5 minutes to complete on CPU with 8 multi-processing threads.

\begin{figure*}
\begin{minipage}{0.33\textwidth}  
	\centerline{
        \subfigure[boot]{
            \includegraphics[width=\linewidth]{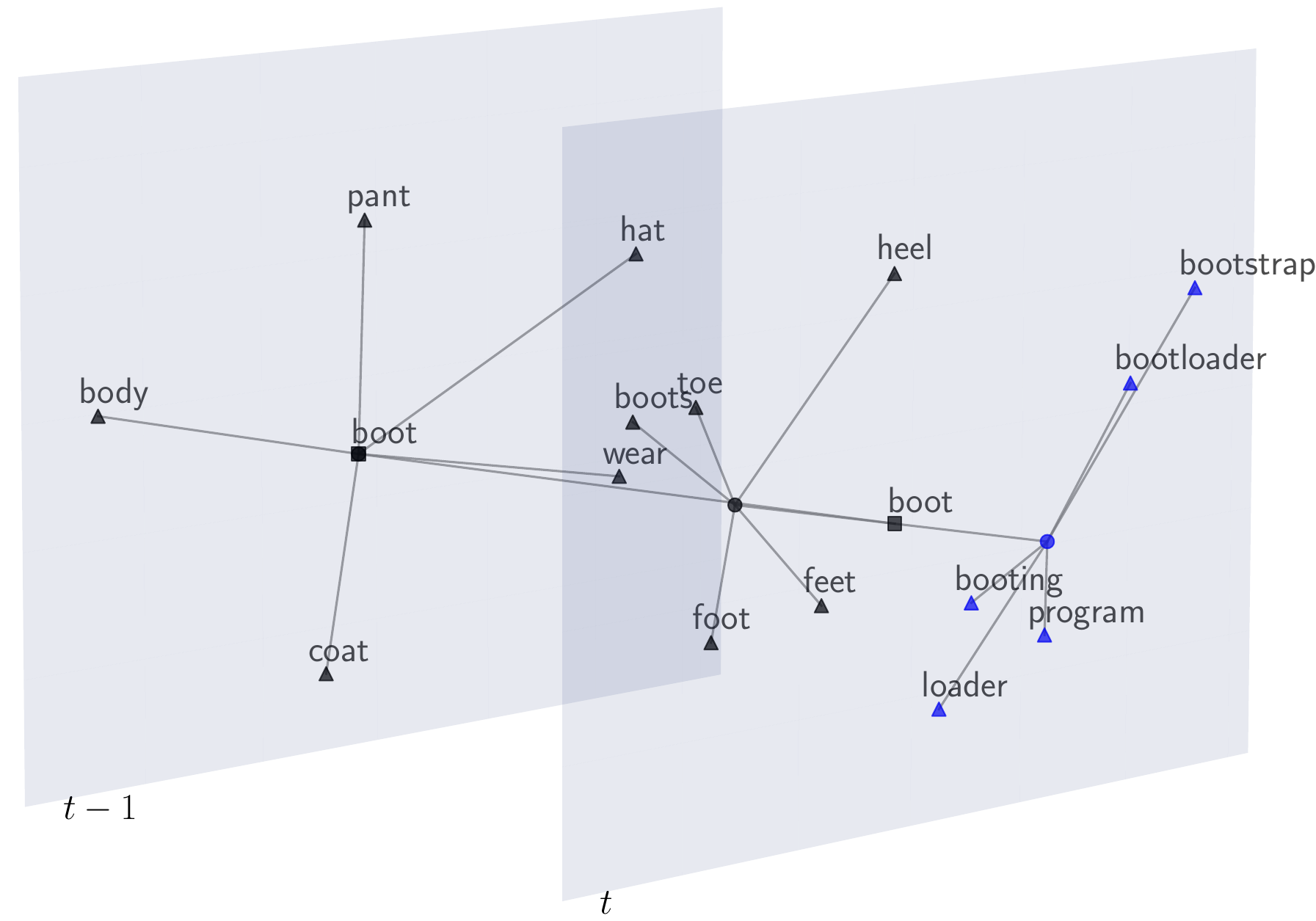}
        }
    } 
\end{minipage} 
\begin{minipage}{0.33\textwidth}  
	\centerline{
        \subfigure[cloud]{
            \includegraphics[width=\linewidth]{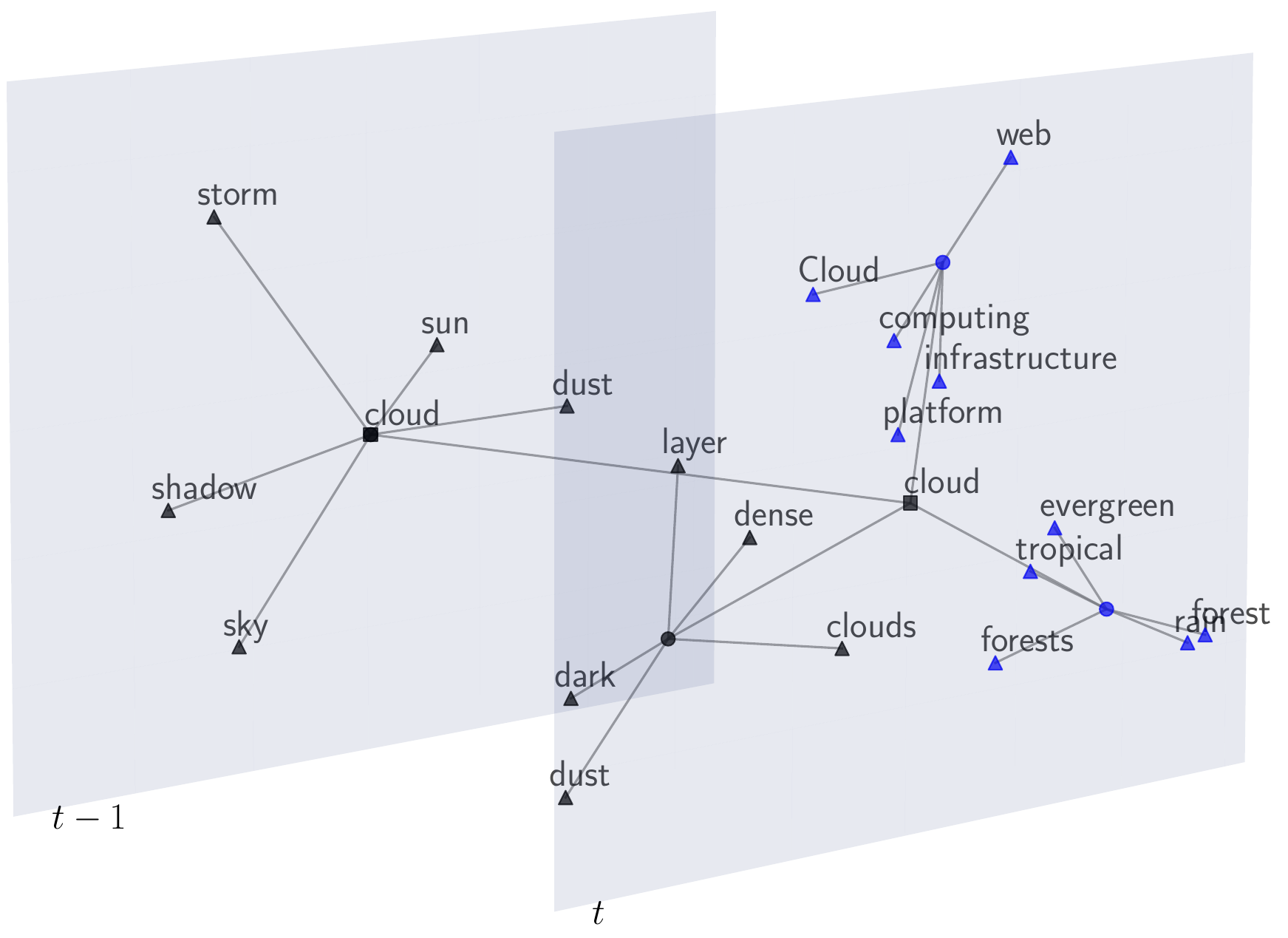}
        }
    } 
\end{minipage} 
\begin{minipage}{0.33\textwidth}  
	\centerline{
        \subfigure[data]{
            \includegraphics[width=\linewidth]{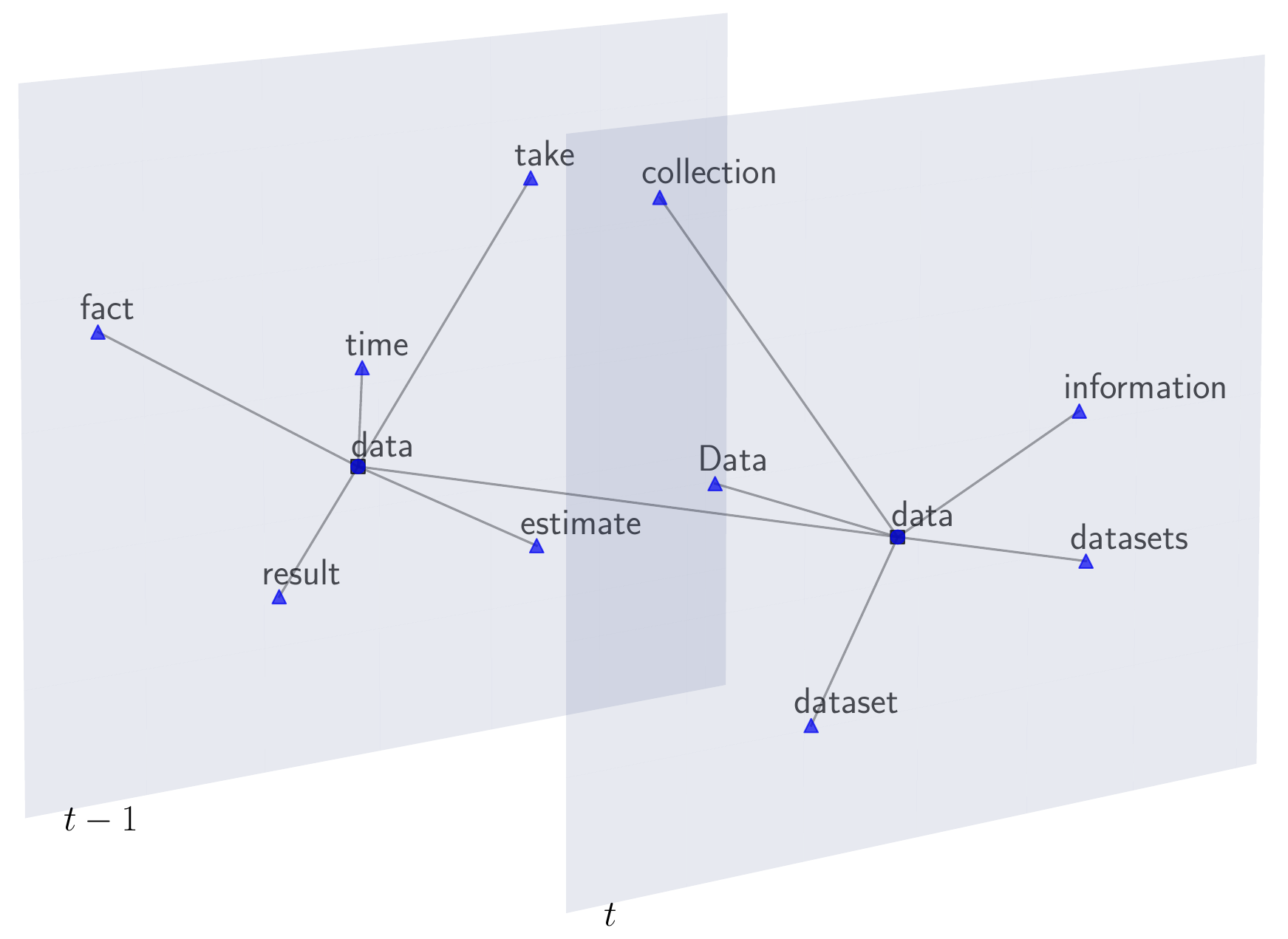}
        }
    }  
\end{minipage} 

\begin{minipage}{0.33\textwidth}  
	\centerline{
        \subfigure[feed]{
            \includegraphics[width=\linewidth]{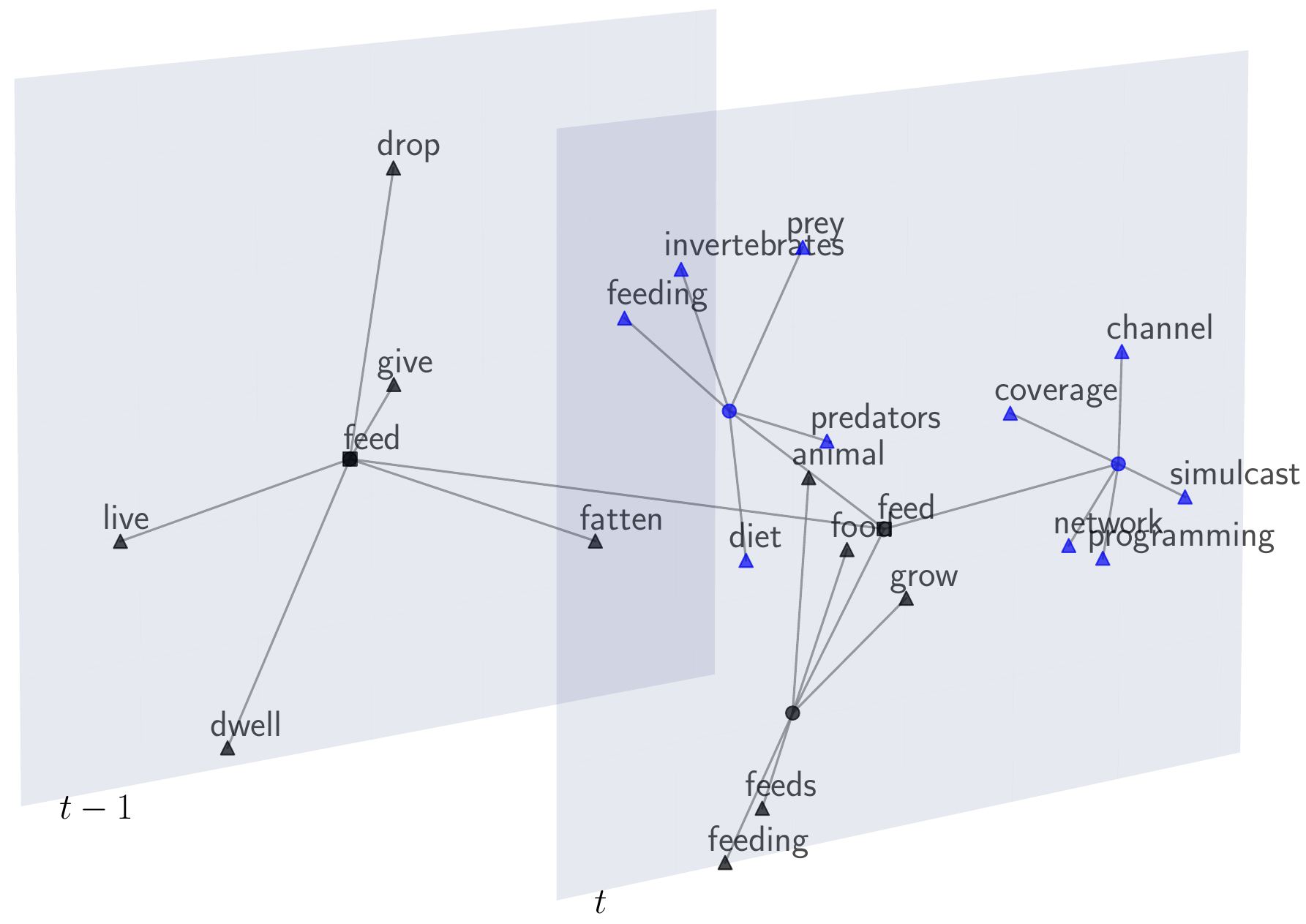}
        }
    } 
\end{minipage} 
\begin{minipage}{0.33\textwidth}  
	\centerline{
        \subfigure[gay]{
            \includegraphics[width=\linewidth]{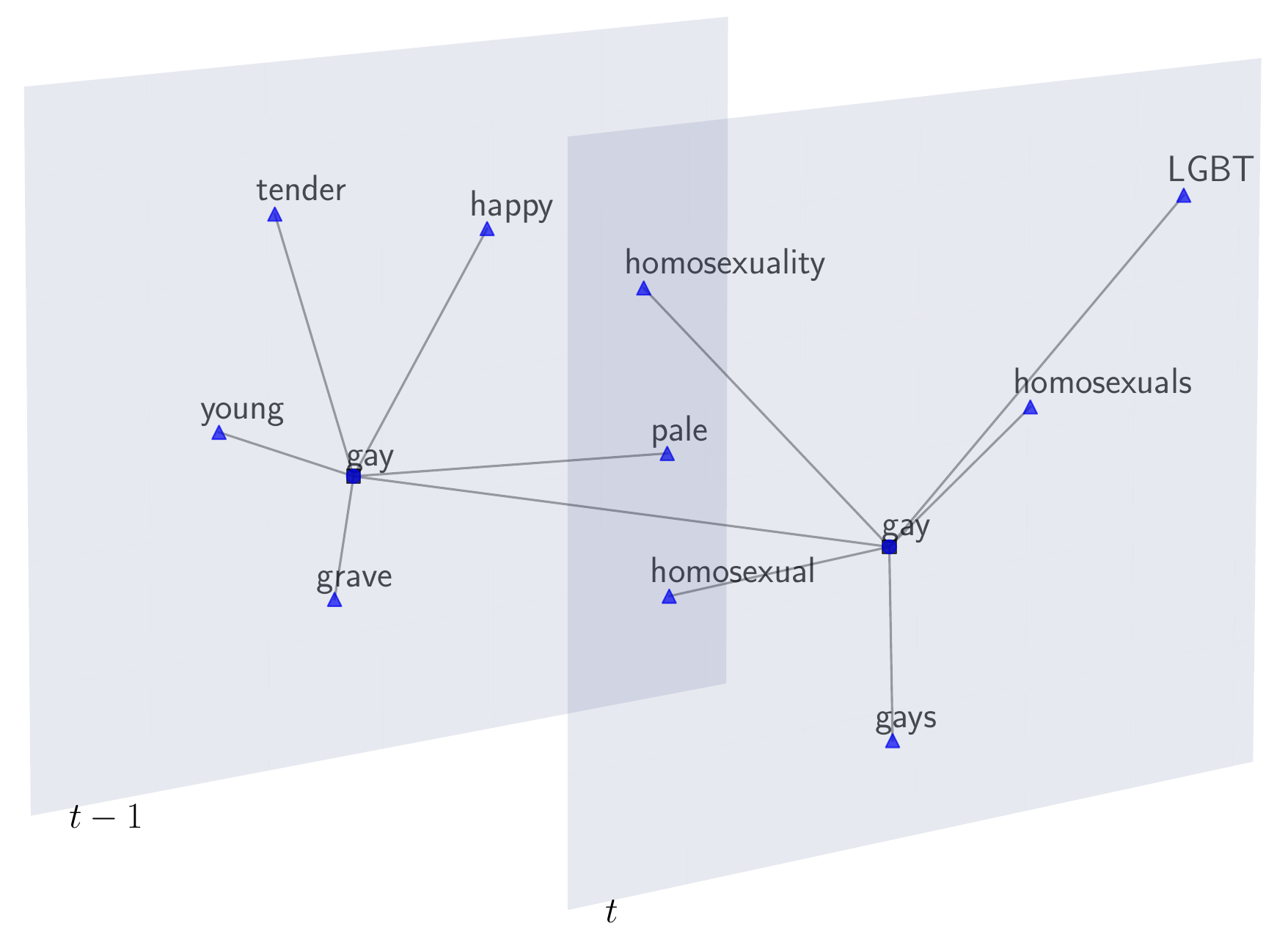}
        }
    } 
\end{minipage} 
\begin{minipage}{0.33\textwidth}  
	\centerline{
        \subfigure[gift]{
            \includegraphics[width=\linewidth]{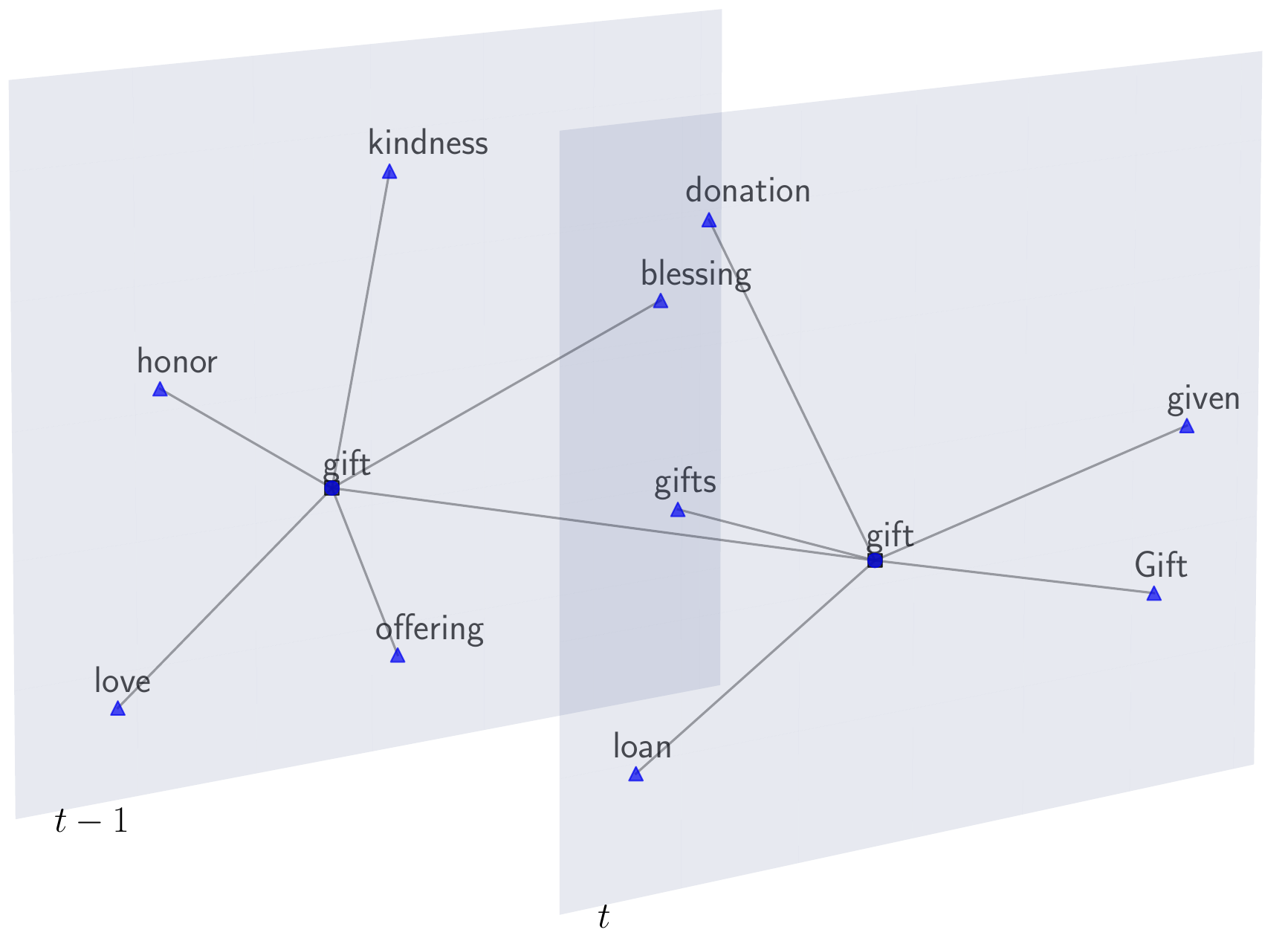}
        }
    }  
\end{minipage} 

\begin{minipage}{0.33\textwidth}  
	\centerline{
        \subfigure[mail]{
            \includegraphics[width=\linewidth]{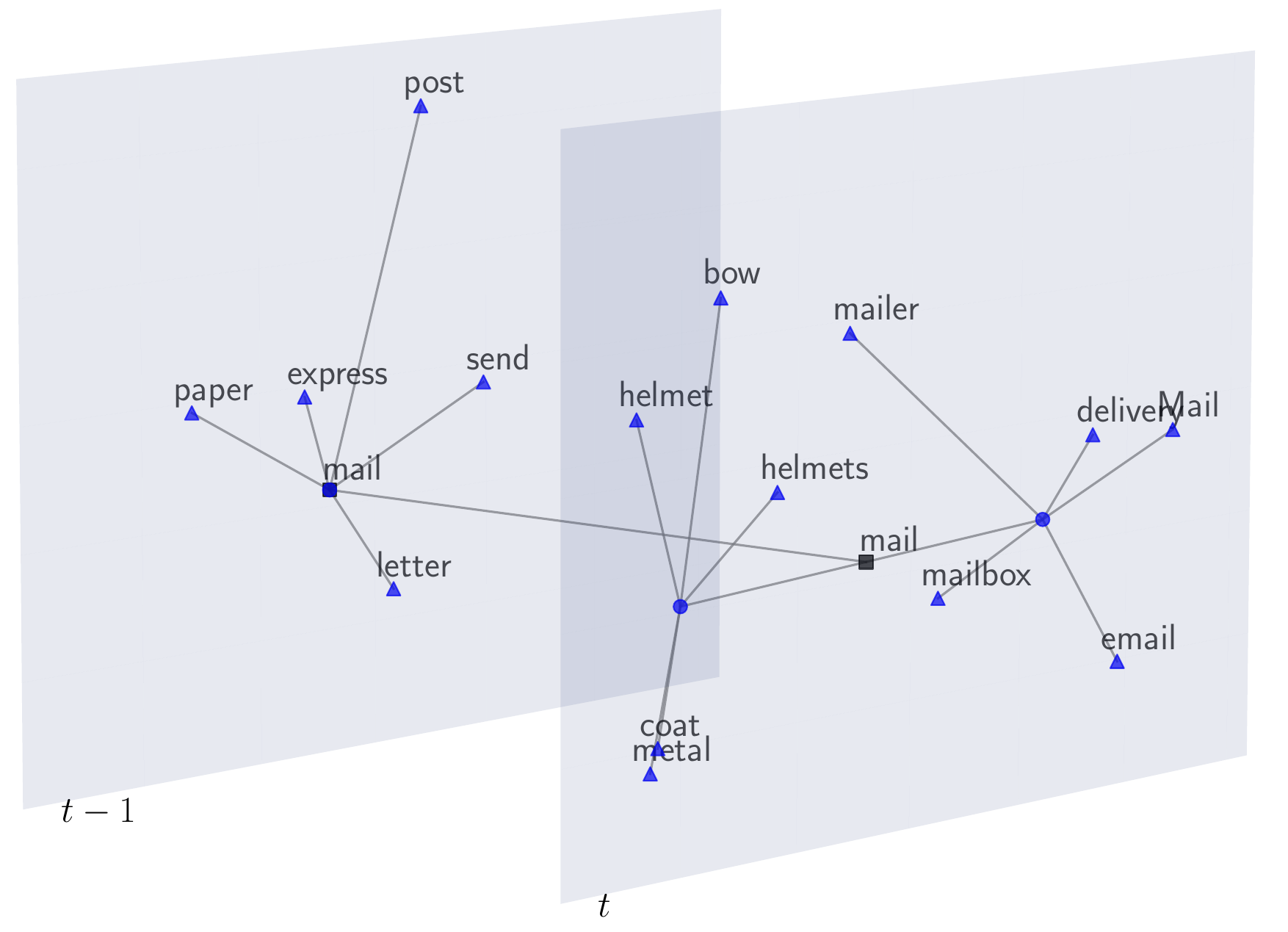}
        }
    } 
\end{minipage} 
\begin{minipage}{0.33\textwidth}  
	\centerline{
        \subfigure[memory]{
            \includegraphics[width=\linewidth]{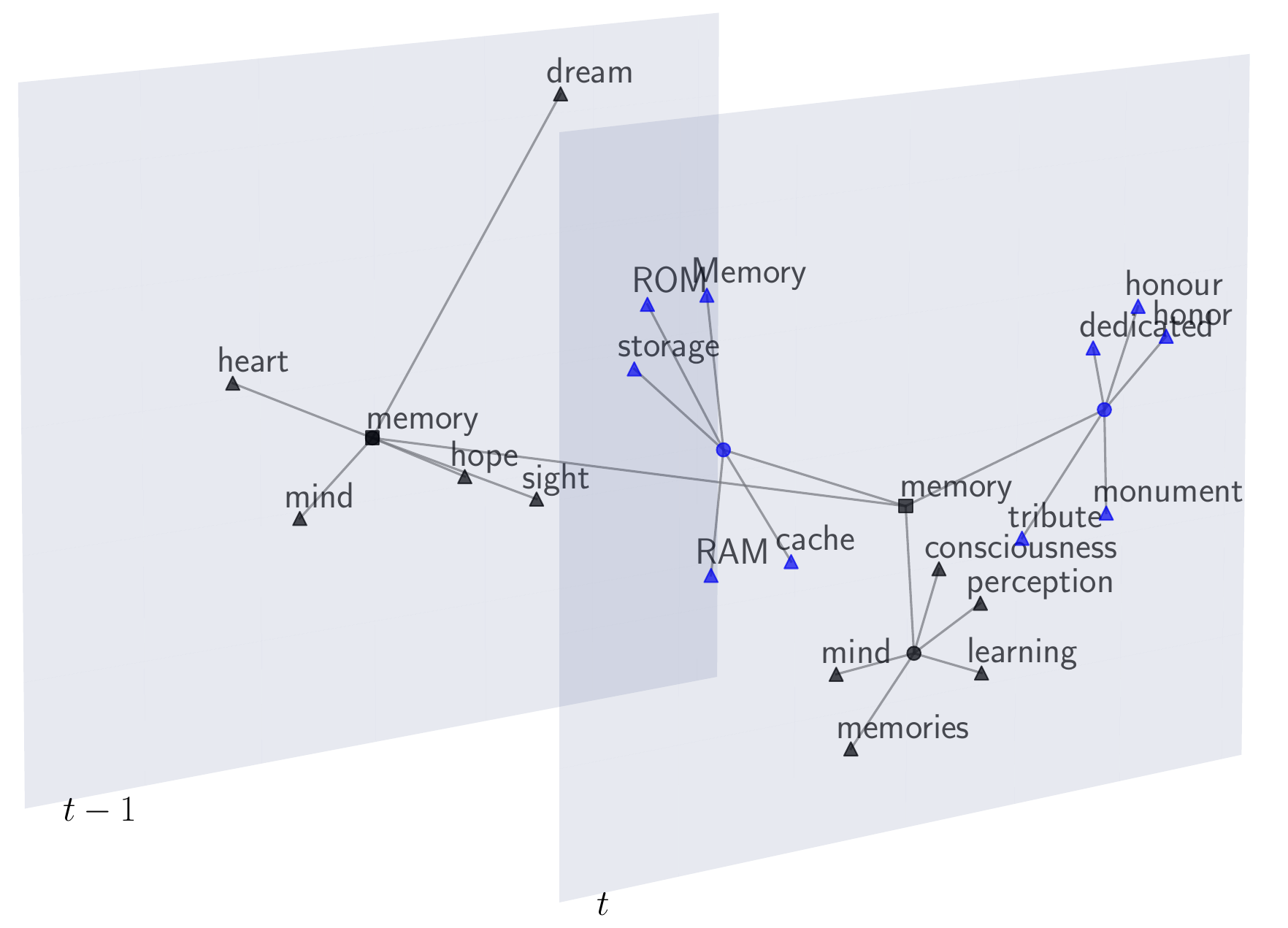}
        }
    } 
\end{minipage} 
\begin{minipage}{0.33\textwidth}  
	\centerline{
        \subfigure[web]{
            \includegraphics[width=\linewidth]{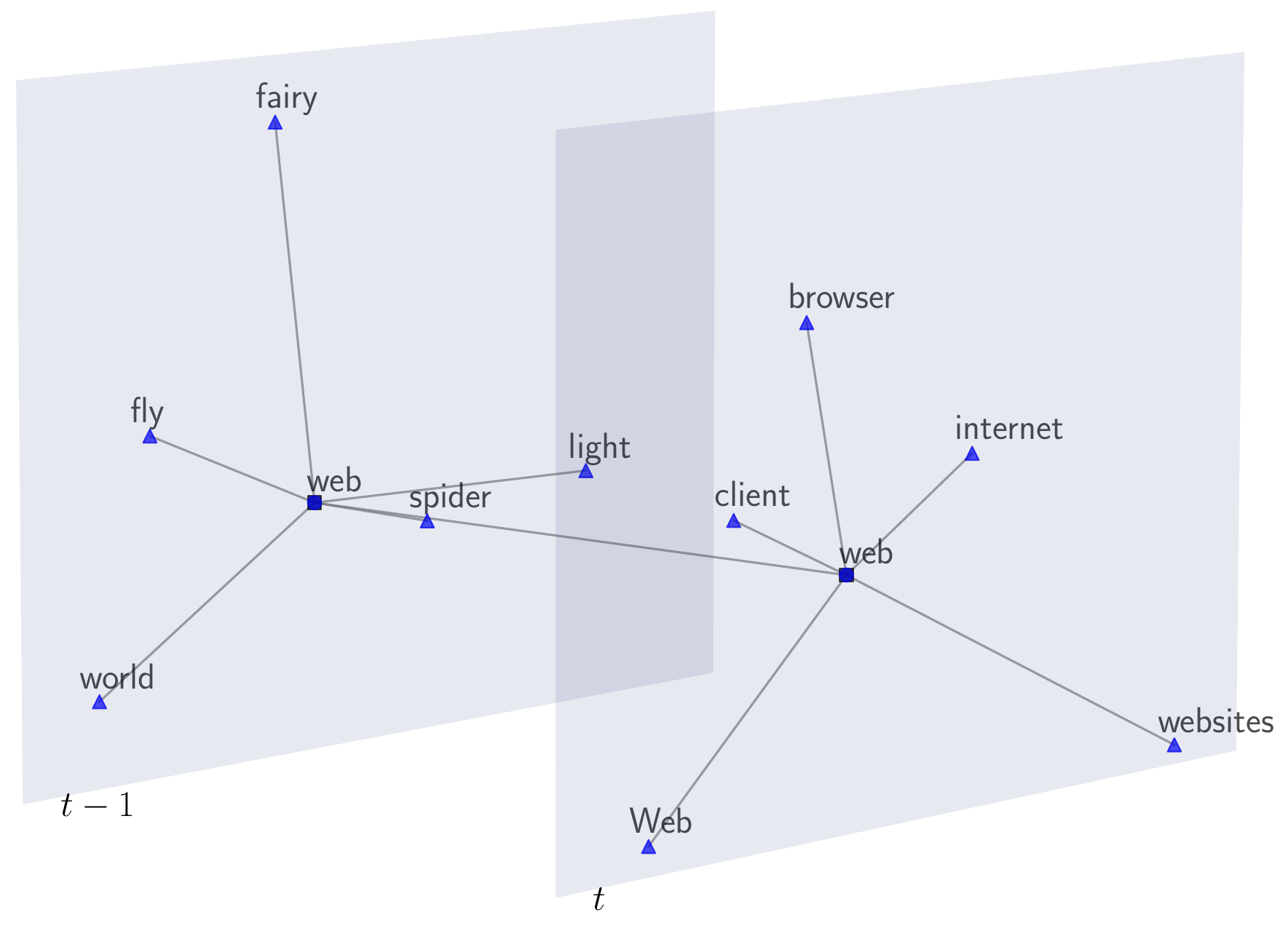}
        }
    }  
\end{minipage} 
\caption{Representation of the detected semantic changes for the English word `boot', `cloud', `data', `feed', `gay', `gift', `mail', `memory', and `web' in our temporal dynamic graph. Blue nodes at time $t$ indicate the acquisition of new meanings, while blue nodes at time $t-1$ indicate the loss of original meanings. Black nodes indicate word meanings that remain unchanged over time.}\label{fig:sc-intralingual-en-appendix}
\end{figure*}

\begin{figure*}
\subfigure[\{memory, Erinnerung, minne\} ]{
\begin{minipage}{0.33\textwidth}  
	\centerline{\includegraphics[width=\linewidth]{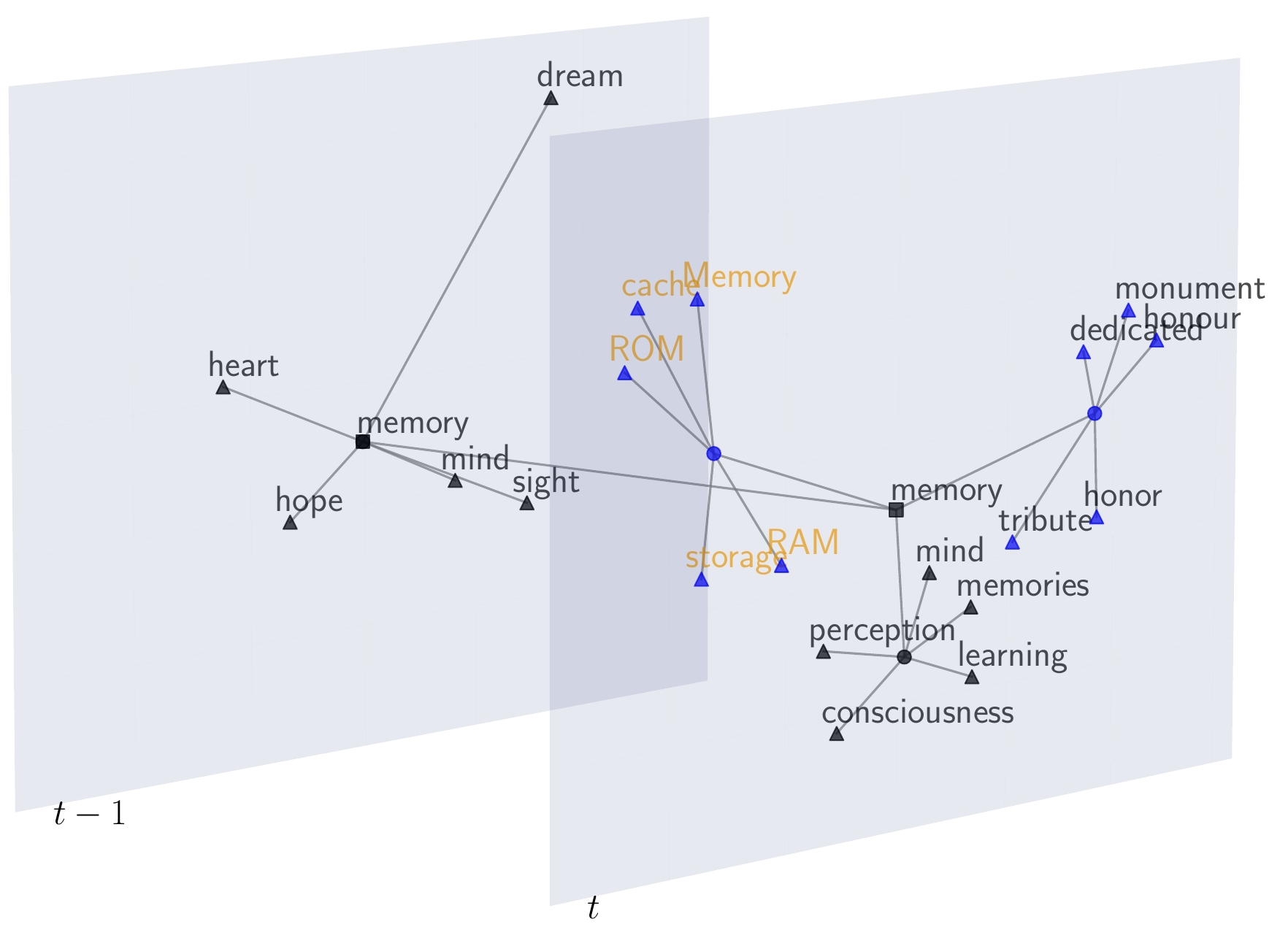}} 
\end{minipage} 
\begin{minipage}{0.33\textwidth}  
	\centerline{\includegraphics[width=\linewidth]{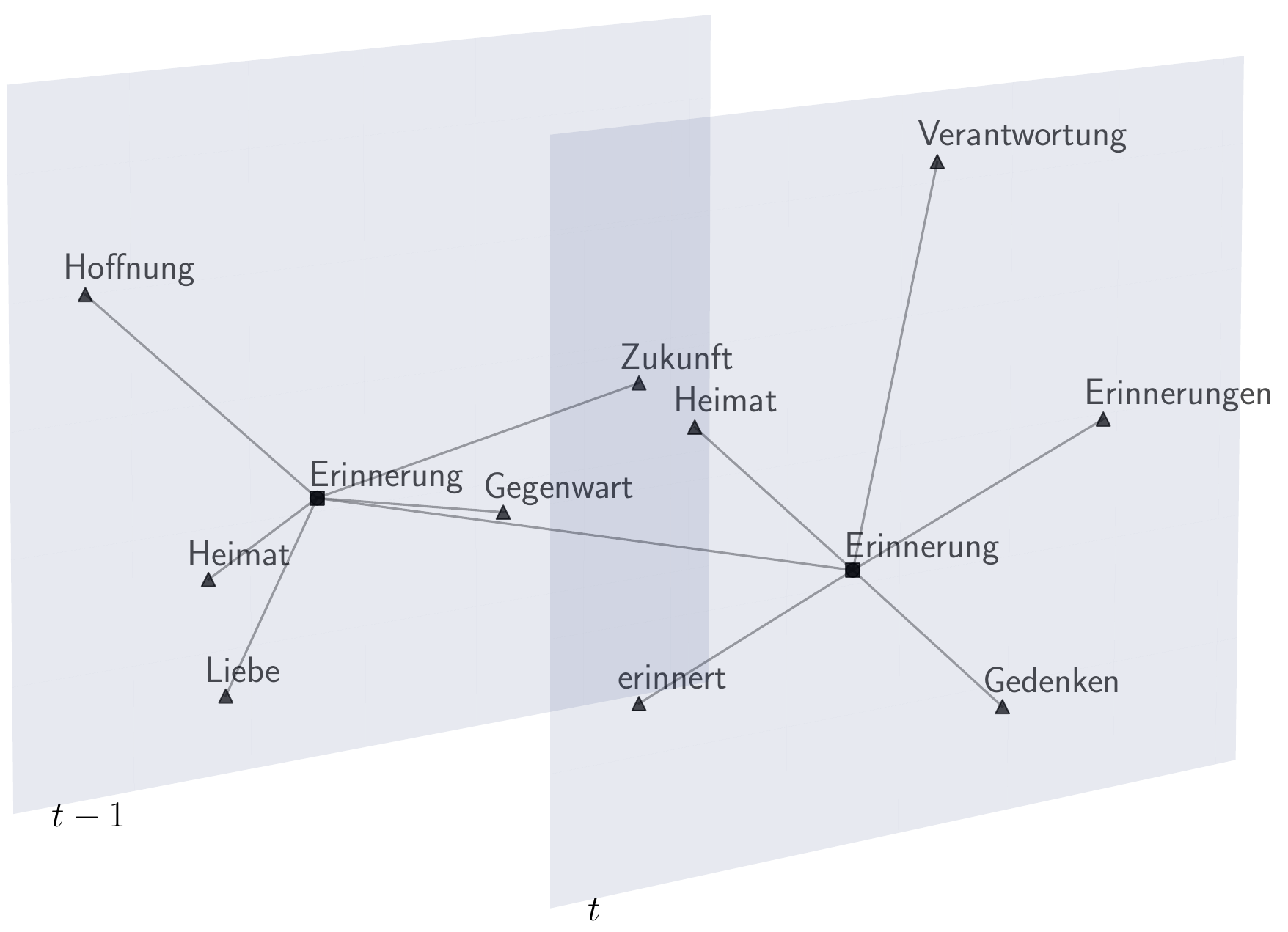}}  
\end{minipage} 
\begin{minipage}{0.33\textwidth}  
	\centerline{\includegraphics[width=\linewidth]{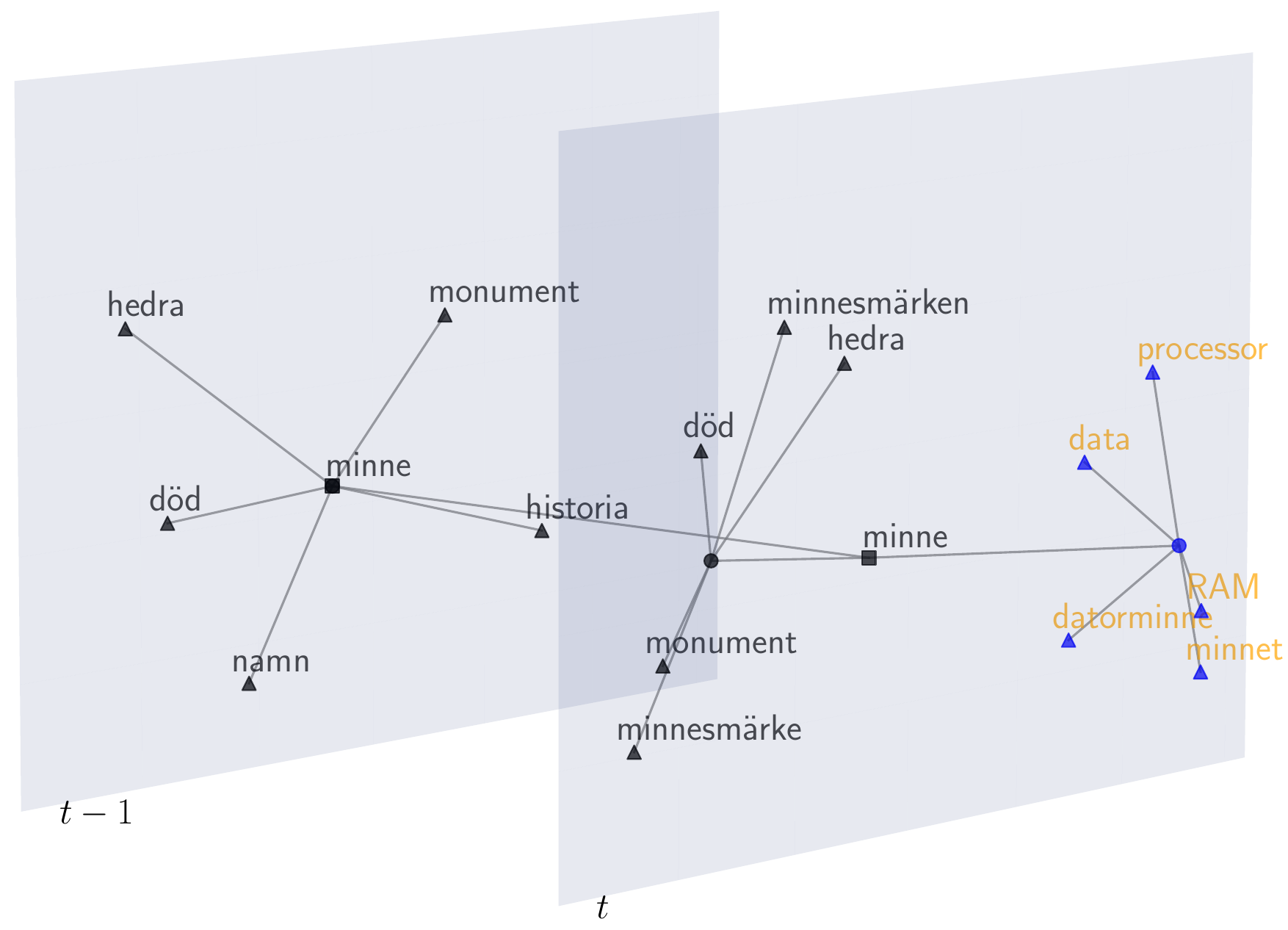}}  
\end{minipage} 
}

\subfigure[\{gay, fröhlich, gay\} ]{
\begin{minipage}{0.33\textwidth}  
	\centerline{\includegraphics[width=\linewidth]{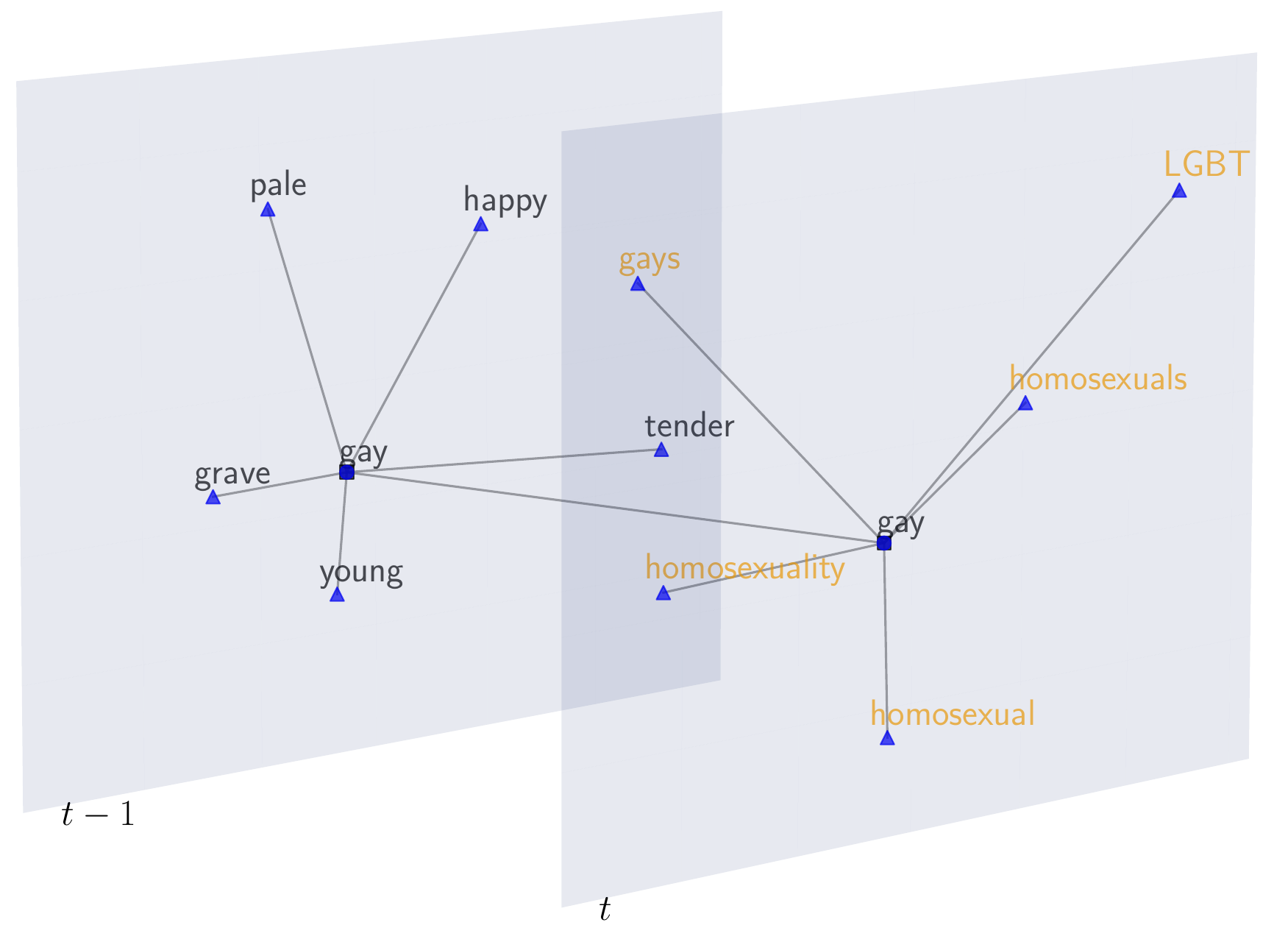}} 
\end{minipage} 
\begin{minipage}{0.33\textwidth}  
	\centerline{\includegraphics[width=\linewidth]{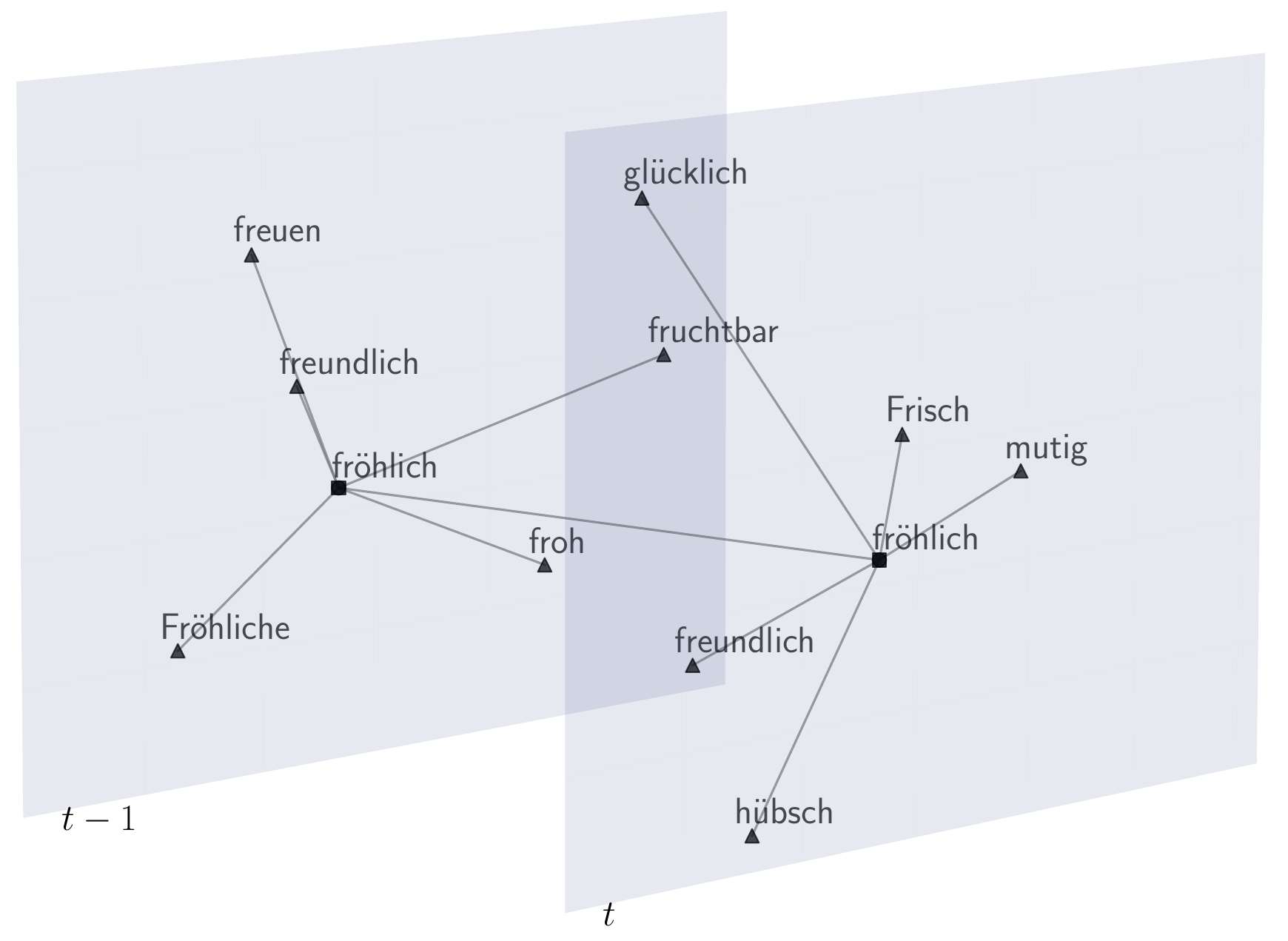}}  
\end{minipage} 
\begin{minipage}{0.33\textwidth}  
	\centerline{\includegraphics[width=\linewidth]{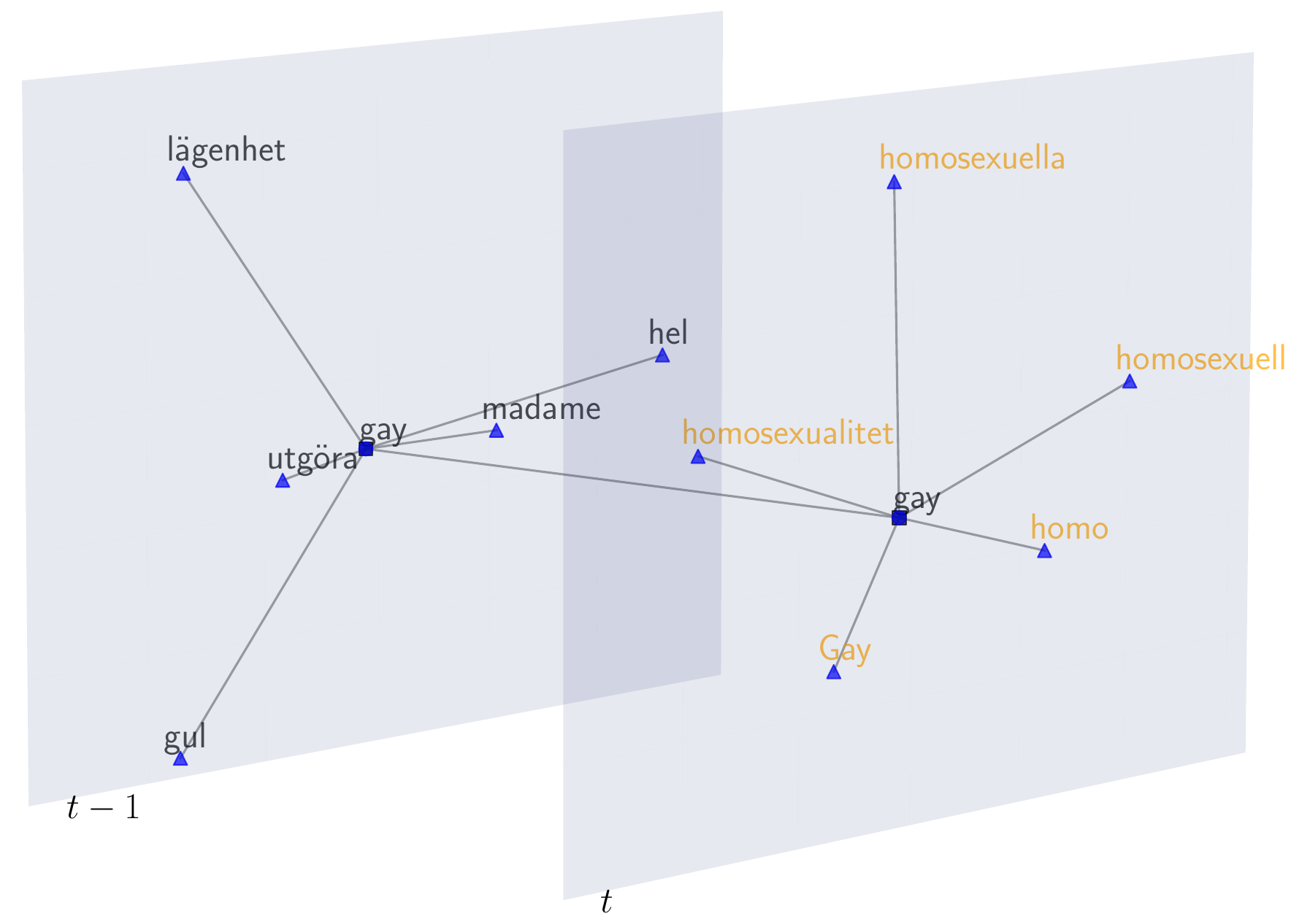}}  
\end{minipage} 
}

\subfigure[\{gift ,Gift, gift\} ]{
\begin{minipage}{0.33\textwidth}  
	\centerline{\includegraphics[width=\linewidth]{pics/en-intra/en-gift.pdf}} 
\end{minipage} 
\begin{minipage}{0.33\textwidth}  
	\centerline{\includegraphics[width=\linewidth]{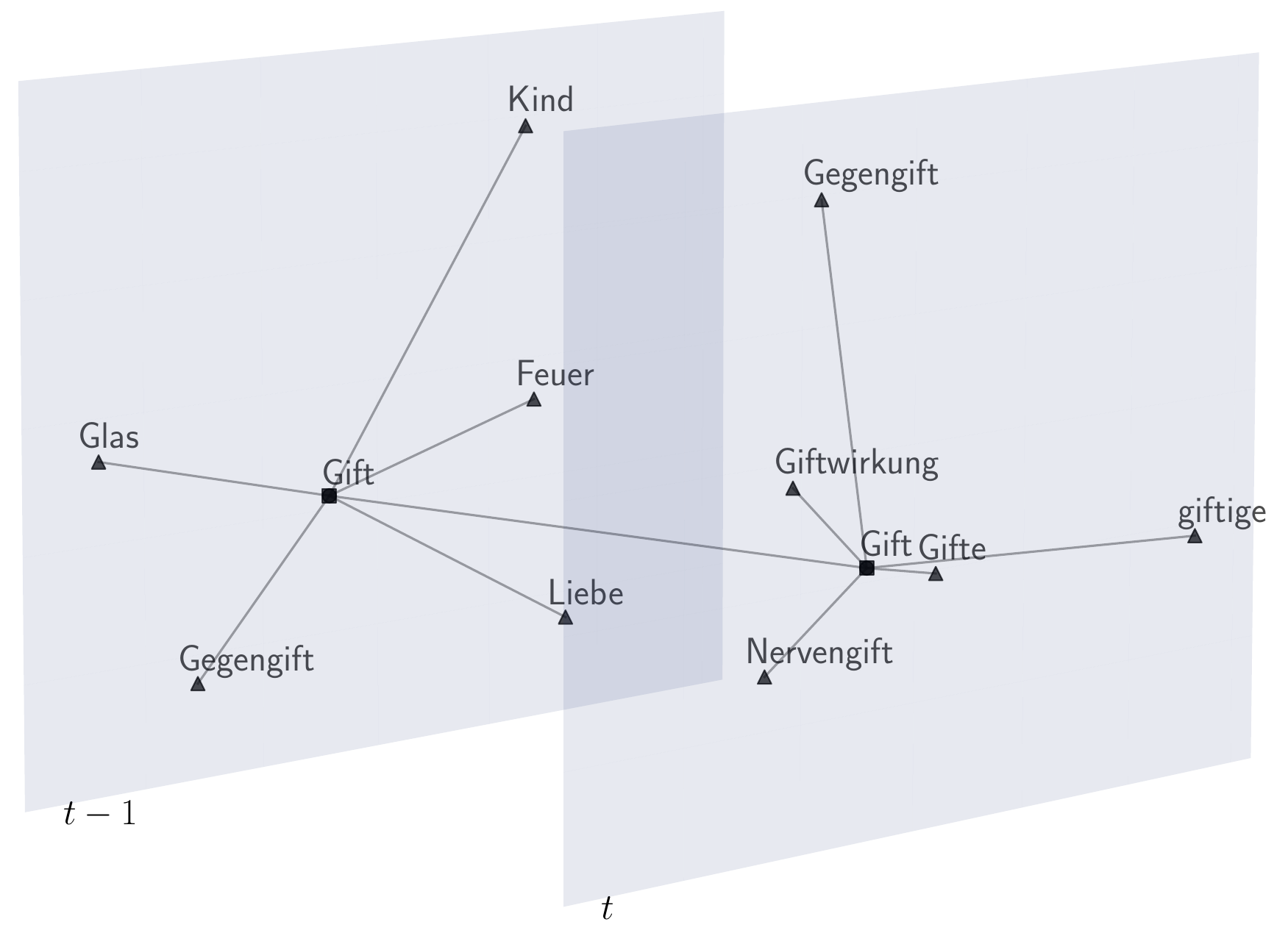}}  
\end{minipage} 
\begin{minipage}{0.33\textwidth}  
	\centerline{\includegraphics[width=\linewidth]{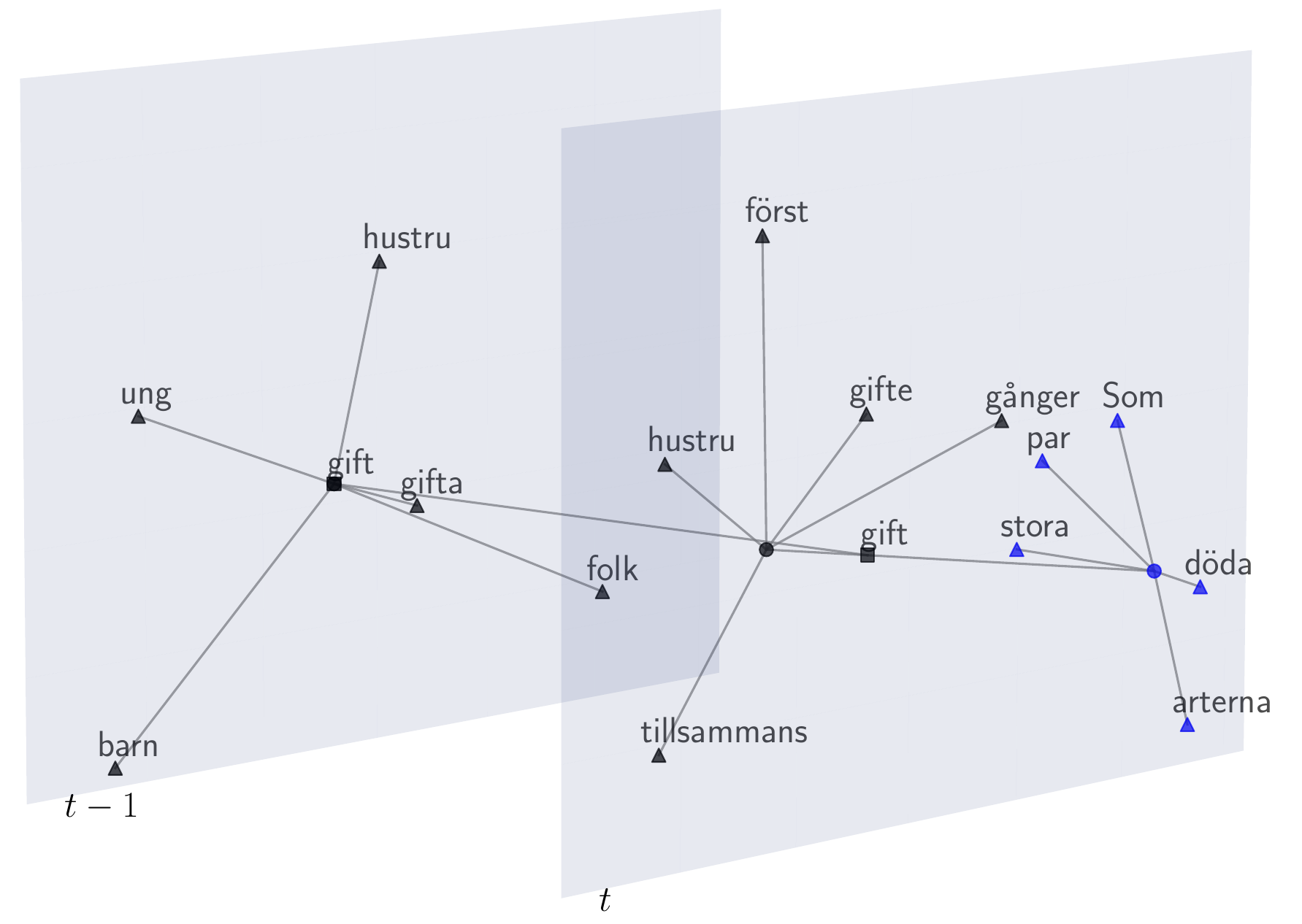}}  
\end{minipage} 
}

\subfigure[\{cloud, Wolke, moln \}]{
\begin{minipage}{0.33\textwidth}  
	\centerline{\includegraphics[width=\linewidth]{pics/en-intra/en-cloud.pdf}} 
\end{minipage} 
\begin{minipage}{0.33\textwidth}  
	\centerline{\includegraphics[width=\linewidth]{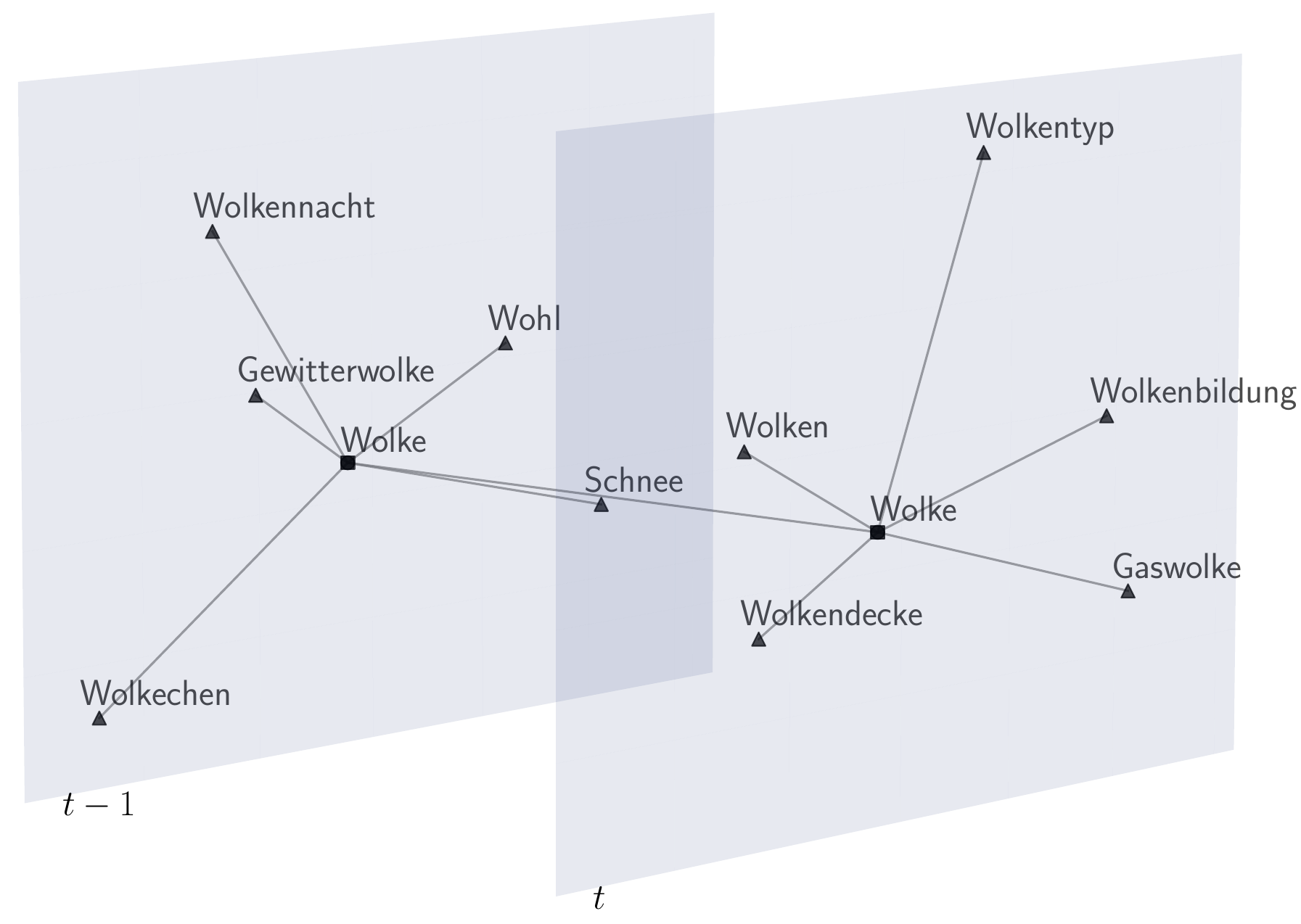}}  
\end{minipage} 
\begin{minipage}{0.33\textwidth}  
	\centerline{\includegraphics[width=\linewidth]{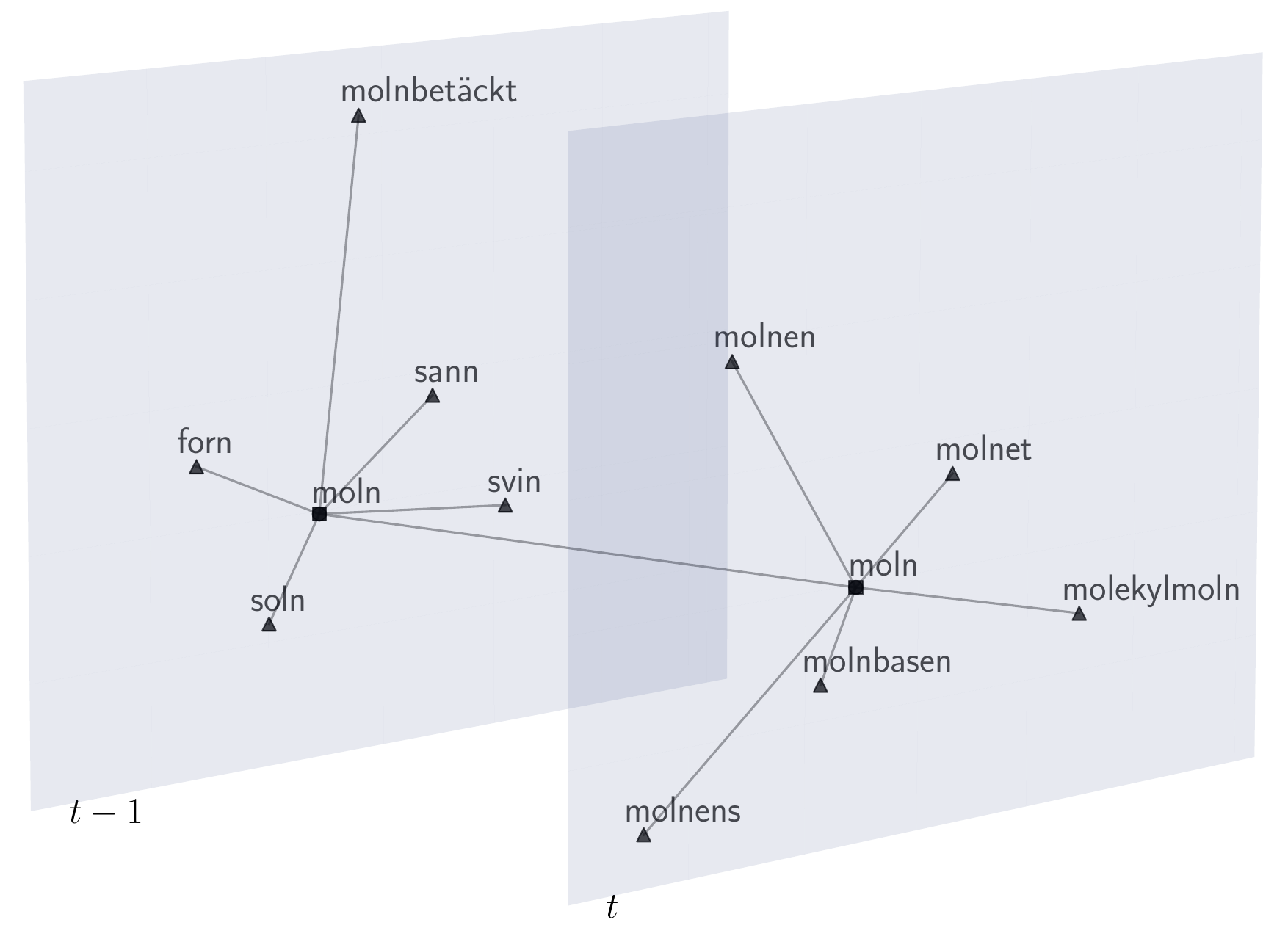}}  
\end{minipage} 
}

\caption{Representation of the detected semantic changes in word translations across English, German and Swedish, shown in our inter-language temporal dynamic graphs. Blue nodes indicate the acquisition of a new meaning at time $t$ (the loss of an existing one at $t-1$), while black nodes indicate meanings unchanged over time. Orange text marks certain changed meanings that are considered highly similar across languages. 
Figure (a) \{memory, Erinnerung, minne\}: `memory' and `minne' gain the same new meaning about computer storage unit at time $t$ while the meaning of `Erinnerung' remains unchanged.
Figure (b) \{gay, fröhlich, gay\}: `gay' (English) and `gay' (Swedish) gain the same new meaning about homosexuality at time $t$, and both words lose their different original meanings that they had at $t-1$. The meaning of `fröhlich' remains unchanged.
Figure (c) \{gift, Gift, gift\}: their semantic changes diverge greatly over time. The meaning of `gift' (English) changes from \textit{a notable act of giving} to \textit{something given voluntarily without payment}. Gift (German) retains the meaning \textit{poison} over time.  `gift' (Swedish) gains a new meaning \textit{poison} perhaps borrowed from German. 
Figure (d) \{cloud, Wolke, moln\}: `cloud' gains a new meaning while the meanings of `Wolke' and `moln' remain unchanged.}\label{fig:sc-xlingual-appendix}
\end{figure*}

\end{document}